%% file: main.tex
\documentclass[11pt]{article}

\usepackage[final]{acl}

\usepackage{silence}
\usepackage{iftex}
\ifluatex
\usepackage{microtype}

\usepackage[english,bidi=default]{babel} % English as the main language.
\babelfont{rm}{TeXGyreTermesX} % similar to Times
\babelprovide[import]{hindi}
\babelfont[*devanagari]{rm}{Lohit Devanagari}
\babelprovide[import]{arabic}
\babelfont[*arabic]{rm}{Noto Sans Arabic}

\else
\usepackage{times}
\usepackage{latexsym}

\usepackage[T1]{fontenc}
\usepackage[utf8]{inputenc}

\usepackage{microtype}

\usepackage{inconsolata}
\fi

\usepackage{graphicx}

\usepackage{amsmath}
\usepackage{booktabs}

\usepackage{tabularx}
\usepackage[table]{xcolor}
\usepackage{array}
\usepackage{amssymb}

\usepackage{makecell}

\newcommand{\RefSpanACES}{\texorpdfstring{{Span-ACES}\textsubscript{Ref}}{Span-ACES-REF}}

\title{Quantifying the Impact of Translation Errors on Multilingual LLM Evaluation}

\author{
  \textbf{Klaudia-Doris Thellmann\textsuperscript{1}},
  \textbf{Bernhard Stadler\textsuperscript{1,2}},
  \textbf{Michael Färber\textsuperscript{1}},
  \textbf{Jens Lehmann\textsuperscript{3,1,2}\thanks{Work done outside of Amazon.}}\\
  {firstname.lastname@tu-dresden.de}
\\
\\
  \textsuperscript{1}TUD Dresden University of Technology and ScaDS.AI,\\
  \textsuperscript{2}InfAI e.V.,
  \textsuperscript{3}Amazon
\\
  \small{
    \textbf{Correspondence:} \href{mailto:klaudia-doris.thellmann@tu-dresden.de}{klaudia-doris.thellmann@tu-dresden.de}
  }
}

\begin{document}
\maketitle
\begin{abstract}
Machine-translated benchmarks are widely used to assess the multilingual capabilities of large language models (LLMs), yet translation errors in these benchmarks remain underexplored, raising concerns about the reliability and comparability of multilingual evaluation.
We address two practical gaps:
(i) how well automatic MQM-style error spans from LLM judges and a span-aware QE baseline (xCOMET-XXL) match expert human span annotations on benchmark translations, and
(ii) how strongly translation errors (as opposed to source-side issues in the English original) explain accuracy drops on translated benchmarks.
We find that span agreement is non-trivial on naturally occurring benchmark translations, and that target-side translation errors are consistently associated with measurable, percentage-point drops in translated accuracy even after controlling for English correctness and source-side anomalies.
\end{abstract}

\section{Introduction}
\label{sec:intro}
In multilingual evaluation, machine-translated datasets are widely used as reference data, yet translation quality is often overlooked, undermining reliability and comparability~\citep{choenni_2024, artetxe_2020, plaza2024}.
Human protocols such as MQM~\citep{lommel_2013, lommel_2024} and Error Span Annotation (ESA)~\citep{kocmi_2024} provide increasingly diagnostic assessments~\citep{freitag_2021}.

More recently, researchers have treated LLMs themselves as translation judges (``LLM-as-a-judge''; \citealp{kocmi_2023a}), using zero- or few-shot prompting to tag MQM-style error spans, while recent fine-grained quality-estimation work has likewise moved toward span-aware error localization, most prominently in xCOMET-XXL~\citealp{guerreiro_2024}; however, its behavior on naturally occurring benchmark translations and the implications for benchmark reliability remain underexplored.

This trend is exemplified by GEMBA and GEMBA-ESA~\citep{freitag_2024}, and by GPT-based evaluators such as AutoMQM~\citep{huang_2024} for inline span detection, or MQM-APE~\citep{lu_2025}, which uses automatic post-editing to refine translations.

Prior work investigating the effects of translation artifacts on model performance relies either on manual inspection of small samples~\citep{artetxe_2020, plaza2024}, which provides qualitative insights but does not scale, or on heuristics~\citep{park_2024, choenni_2024} such as sentence length ratios or learned quality estimation scores (e.g., COMET-QE~\citep{rei_2020}), both of which lack precision in identifying the type and location of translation errors.
In addition, most of these studies are limited to single benchmarks or languages (e.g.~Spanish MMLU~\citep{plaza2024} or XNLI~\citep{artetxe_2020}).

\definecolor{srcband}{HTML}{EAF2FB}
\definecolor{tgtband}{HTML}{FBEFE6}
\newcolumntype{Y}{>{\raggedright\arraybackslash}X}

\begin{table*}[tb]
\centering
\footnotesize
\setlength{\tabcolsep}{3pt}
\renewcommand{\arraystretch}{1.00}
\begin{tabularx}{\textwidth}{l l l Y}
\toprule
\textbf{Source} & \textbf{EN key phrase} & \textbf{Translation key phrase} & \textbf{Why it matters for evaluation} \\
\midrule
\multicolumn{4}{@{}l}{\textit{Source-side anomalies (S) -- flagged on the English source}} \\
\rowcolor{srcband}ARC (DA), corrected & rate of speed      & samme hastighed        & Source oddness corrected in target; not a target-side translation error. \\
\rowcolor{srcband}ARC (DE), preserved & terrestrial plants & terrestrische Pflanzen & Source anomaly (plants instead of planets) preserved in target; may be mistaken for a target-side error. \\[2pt]
\multicolumn{4}{@{}l}{\textit{Target-side translation errors (T) -- English solvable, translation not}} \\
\rowcolor{tgtband}MMLU (FR)      & Wrong             & Faux                      & Moral judgment $\rightarrow$ factual judgment. EN\,\checkmark\,/\,FR\,$\times$ \\
\rowcolor{tgtband}ARC (DA)       & shorter distance  & kortere tid               & Stopping distance $\rightarrow$ stopping time. EN\,\checkmark\,/\,DA\,$\times$ \\
\rowcolor{tgtband}GSM8K (SL)     & feet / cubic feet & metra / kubi\v{c}na metra & Unit shift invalidates the calculation. EN\,\checkmark\,/\,SL\,$\times$ \\
\rowcolor{tgtband}HellaSwag (DA) & liner             & foderet                   & Cleaning action $\rightarrow$ feeding action. EN\,\checkmark\,/\,DA\,$\times$ \\
\bottomrule
\end{tabularx}
\caption{Illustrative examples of source-side anomalies (S) and target-side translation errors (T). S cases are flagged on the English source and either corrected or preserved in the translation; they do \emph{not} count as target-side translation errors. T cases change the intended meaning of the target item; in our sample, the English original was solved correctly (EN\,\checkmark) while the translation was not (target\,$\times$).}
\label{tab:examples_st}
\end{table*}

Table~\ref{tab:examples_st} illustrates why source-side anomalies matter: some are corrected in translation, while others are preserved or exacerbated, so source- and target-side errors can independently affect both benchmark outcomes and the construct being tested.

Two practical questions remain under-addressed: (i) how well do LLM-produced MQM-style spans match expert human spans on real benchmark translations, and (ii) how strongly do translation errors (vs.\ source-side issues) explain accuracy drops on translated benchmarks?

We study these questions on the EU20 suite~\citep{thellmann_2024} and annotate translation errors with four LLM annotators using an MQM-inspired prompt (\S\ref{sec:tqann}).
To strengthen the comparison, we also include xCOMET-XXL as a baseline, complement thresholded span matching with threshold-free character-overlap analyses, and test in a counterfactual analysis whether removing translation-error effects changes downstream system rankings.

We make three contributions:
\paragraph{1. Human reference and LLM span agreement on EU20 (\S\ref{sec:tqann}).}
We release a professional span-level MQM reference for an EU20 subset (225 items, nine target languages) and evaluate four LLM annotators and xCOMET-XXL against it.
GPT-5.2 shows the highest agreement with human annotations under strict span matching (mean span-level F1 \textbf{0.55} under position-overlap matching) and complementary character-overlap metrics (Char-F1 / Char-F1w).

\paragraph{2. Meta-evaluation with \RefSpanACES{} (\S\ref{sec:RefSpanACES}).}
We evaluate span localization on \RefSpanACES{} (1{,}407 items), a cleaned projection of SPAN-ACES~\citep{moghe_2025}, and validate the transformation on 178 records where manual review and GPT-5.2 agree on 165/178 (0.93), enabling controlled classic vs.\ tolerant span metrics.
\paragraph{3. Performance impact with source controls (\S\ref{sec:tqperf}).}
Using our annotation-and-evaluation setup, we estimate the impact of target-side translation errors ($T$) and source-side issues ($S$) on translated accuracy via logistic regression (fixed effects; English controls; bootstrap CIs): across annotators,
$T$ is associated with drops of about \textbf{6--8 pp} (full model) and \textbf{6--11 pp} among English-solvable items.
A counterfactual $T{=}0$ ranking analysis further shows that translation errors mainly shift absolute scores rather than relative system rankings.

The annotated datasets, \RefSpanACES{} resources, and our codebases and prompts are available under \href{https://hf.co/btqe}{hf.co/btqe} and \href{https://github.com/btqe}{github.com/btqe}.

\section{Related Work}
\label{sec:rel_work}
\paragraph{Translation artifacts and their effects.}
Several studies have shown that translation artifacts can undermine the reliability of model evaluation:
\citet{choenni_2024} found that MT-generated test sets may overestimate model capabilities, especially in low-resource languages; \citet{artetxe_2020} demonstrated that subtle ``translationese'' can bias cross-lingual benchmarks like XNLI; \citet{plaza2024} reported that mistranslations in Spanish MMLU data cause 6--13\% accuracy loss for GPT-4, with up to 60\% of failures directly linked to translation errors; and \citet{park_2024} observed similar effects for VQA models.
Agrawal et al.~\citep{agrawal-etal-2024-translation} further show that translation errors can substantially affect cross-lingual learning outcomes, especially in lower-resource settings, reinforcing that benchmark translation quality is not merely a surface issue but can alter downstream conclusions.
While these findings underscore the need for rigorous quality control, prior work remains limited in scale and granularity.
Quality estimation and automated translation evaluation have also moved toward span-aware error localization, for example with xCOMET-XXL~\citep{guerreiro_2024} and recent WMT shared tasks on fine-grained evaluation~\citep{blain-etal-2023-findings,zerva-etal-2024-findings,lavie-etal-2025-findings}.

Our work differs in focusing on naturally occurring benchmark translations and on linking localized translation errors to downstream benchmark accuracy.

\paragraph{Multilingual benchmarks.}
Recent multilingual benchmarks range from carefully curated, manually translated datasets (e.g., SuperGLEBer~\citep{pfister_2024}, ScandEval~\citep{nielsen_2023},  IberoBench~\citep{baucells_2025}, FrenchBench (introduced as part of CroissantLLM; \citealp{faysse_2025}), BenCzechMark~\citep{fajcik_2025}) to large-scale resources generated via machine translation.
While they offer high quality, manually constructed benchmarks are costly and difficult to scale, prompting the use of machine translation for broader coverage, e.g. Global MMLU~\citep{singh_2025}, XNLI~\citep{conneau_2018}, OKAPI~\citep{lai_2023a}, and LAMBADA~\citep{paperno_2016}.
However, many such resources lack transparent quality control.
%Our work advances the field by combining automated span-level error annotation with statistical analysis to assess translation error impact on model performance.
Our work advances this line of research by combining automated span-level error annotation, benchmarked against expert human annotations and a strong baseline, with statistical analysis to assess how translation errors affect downstream LLM performance on multilingual benchmarks.

\section{Human and LLM MQM Annotation}
\label{sec:tqann}
In this section, we present our MQM-based annotation setup for capturing translation errors and assess the span-level annotation accuracy of LLM-based annotators.
We compare LLM-generated MQM annotations against a professionally produced human reference on an EU20 subset and additionally evaluate LLM annotators on \RefSpanACES.

\subsection{Methodology}
\label{sec:tqann:methodology}

\paragraph{Human Annotation.}
Using EU20~\citep{thellmann_2024} (DeepL translations\footnote{\href{https://developers.deepl.com/docs}{developers.deepl.com/docs}} of MMLU~\citep{hendrycks_2021}, ARC~\citep{clark_2018}, HellaSwag~\citep{zellers_2019}, GSM8K~\citep{cobbe_2021}, and TruthfulQA~\citep{lin_2022} into 20 official EU languages), we create a span-level human reference for translation error analysis by manually annotating nine target languages spanning Germanic (DE, DA), Romance (FR, IT, RO), Uralic (ET, HU), Slavic (SL), and Baltic (LT), covering a range of high- to low-resource settings.

For each language, we randomly selected 25 segments (225 total) using fixed per-dataset quotas to balance task types; a minimum-length filter excluded trivially short instances.
Dataset statistics are reported in Appendix~\ref{sec:app:tqann:guide} (Table~\ref{tab:subset-stats}).

Annotation was carried out by professional translators/linguists with one annotator per language using an MQM-inspired, span-based protocol implemented in a custom Argilla interface (Appendix~\ref{sec:app:tqann:guide}).
For segments with errors, annotators marked erroneous spans and assigned an MQM label and severity (Major/Minor), distinguishing \textit{Accuracy} errors (faithfulness/meaning transfer; Table~\ref{tab:mqm-labels-accuracy}) from \textit{Fluency/Style} errors (grammaticality and naturalness; Table~\ref{tab:mqm-labels-fluency}).
They highlighted target-side error spans and, where applicable, the corresponding source spans (Figure~\ref{fig:argilla-ui-annot}), and then submitted a minimally post-edited corrected translation and optional clarification comment (Figure~\ref{fig:argilla-ui-postedit}).

After collection, we performed systematic consistency checks (e.g., presence of error labels/severity, and source--target span alignment where applicable) and investigated irregular cases such as source--target span mismatches by consulting annotator comments and the original segment context.
Where necessary, we corrected format inconsistencies and obvious annotation-entry issues to obtain a clean, machine-readable reference.
In addition, we conducted a targeted manual cross-check by comparing human-annotated spans with LLM-proposed annotations as a complementary sanity check.

\paragraph{LLM-based Annotation.}
To generate automatic, span-level MQM annotations, we use four instruction-tuned LLMs: GPT-5.2, GPT-4o-mini, LLaMA-4, and Mistral-Large. Model references are listed in Appendix~\ref{sec:app:tqann:guide} (Table~\ref{tab:annotator-models}).
For comparison, we additionally include xCOMET-XXL (src+mt, reference-free/QE) as a strong non-generative baseline with span-level error predictions.
We adopt a variant of the GEMBA-ESA prompting approach\footnote{\href{https://github.com/MicrosoftTranslator/GEMBA}{github.com/MicrosoftTranslator/GEMBA}} and extend it with curated multilingual few-shot examples, covering a broad range of error types and both structured content (e.g., multiple-choice questions) and general-purpose text.
Our structured JSON output also aligns with the broader move toward explicit span-tagging formats for translation error detection in reasoning-sensitive settings, such as tagged span annotation~\citep{yeom-etal-2025-tagged}.
To ensure comparability with the reference, the prompt restricts labels to the same MQM inventory (Accuracy vs.~Fluency/Style, see Table~\ref{tab:mqm-labels-accuracy} and~\ref{tab:mqm-labels-fluency} from Appendix~\ref{sec:app:tqann:guide}).
Each model is prompted to produce a JSON-structured response containing a list of annotated error spans with MQM label and severity.

In addition, we annotate source-side anomalies on the same EU20 subset using GPT-5.2 to account for cases where the source has irregularities that may affect downstream TQE.
These annotations follow an MQM-inspired, span-based protocol for fluency/coherence anomalies and distinguish surface issues (e.g., \textit{typo}, \textit{grammar}, \textit{punctuation}, \textit{awkward}) from semantic oddness (e.g., \textit{contradiction}, \textit{broken\_coreference}, \textit{implausible\_logic}), with severity (major/minor) indicating potential impact on interpretability or answerability.

\paragraph{Metrics for Span-Level Agreement.}
Because span annotation is a unitizing task---annotators define both the presence and the boundaries of spans, rather than labeling pre-segmented units---we complement pairwise span-matching scores with reliability analyses that serve as robustness checks, including Krippendorff's unitized alpha ($\alpha_u$) for free-span annotation settings~\citep{krippendorff-1995-unitizing,krippendorff-etal-2016-unitizing}.
This unitizing setting is also reflected in NLP annotation tooling such as DKPro Agreement~\citep{meyer-etal-2014-dkpro}.

We report span-level Recall and F1 computed with greedy 1:1 matching with fixed thresholds and micro-aggregation over items (TP/FP/FN summed per language).
For the position-based overlap coefficient (OC), a candidate pair is matchable if $\mathrm{OC} \ge 0.8$, where $\mathrm{OC}$ is overlap length divided by minimum span length.
For the string-based metric (SIM), we use raw character 3-gram Dice similarity (no normalization) with threshold $\mathrm{SIM} \ge 0.6$.
SIM is computed on spans deduplicated by exact target-span text (on both sides), whereas OC operates on the full span lists and requires valid target offsets.
Spans with missing offsets cannot be matched under OC and therefore contribute to FP/FN in the OC-based Span-F1 computation.

Beyond thresholded OC/SIM matching, we also report threshold-free character-overlap metrics.
Our severity-aware character-overlap metric Char-F1w follows the overlap logic of the WMT25 span-level evaluation setting, giving full credit to severity-matched overlaps and partial credit to severity mismatches~\citep{lavie-etal-2025-findings}.
We additionally report a binary Char-F1 variant that ignores severity labels.
These character-level views complement the thresholded span-matching scores used in the main text and help test whether annotator ranking is robust to metric choice.

The SIM-based results are reported in Appendix~\ref{sec:app:tqann:matching} (Table~\ref{tab:agreement-sim}).
Additional unitizing-reliability analyses are reported in Appendix~\ref{sec:app:tqann:reliability} (Table~\ref{tab:reliability-main}); threshold-free character-overlap results are reported in Appendix~\ref{sec:app:tqann:charoverlap} (Table~\ref{tab:charoverlap-main}), with complementary threshold-sweep ranges in Table~\ref{tab:threshold-sweep}.

\paragraph{Source-overlap.}
To quantify the tendency to annotate source-side oddness as target errors, we compute the fraction of an annotator's target spans (with valid offsets) that overlap target regions linked to GPT-5.2 source-anomaly annotations (OC threshold $0.8$, each annotator span counted at most once).
For each source anomaly, GPT-5.2 provides an English anchor span and, when possible, a target-side anchor in the translation.
We ground these anchors conservatively: provided offsets are validated by exact substring checks, and if offsets are missing or don't match, we attempt exact substring search to recover offsets only when the match is unambiguous.
Otherwise, the record remains unlinked and is excluded from offset-based computations.
We then compute source-overlap with OC threshold 0.8, counting each annotator span at most once via its best overlap.

\subsection{Results and Discussion}
\label{sec:tqann:results}

\begin{table}[thb]
\centering
\small
\setlength{\tabcolsep}{3pt}
\begin{tabular}{l c c c c}
\toprule
\textbf{Annotator}
  & \makecell{\textbf{OC}\\\textbf{Recall}}
  & \makecell{\textbf{OC}\\\textbf{$F_1$}}
  & \makecell{\textbf{$F_1$ range}\\\textbf{(lang.)}}
  & \makecell{\textbf{Char-$F_1$}\\\textbf{{[}95\% CI{]}}} \\
\midrule
GPT-5.2               & .44 & \textbf{.55} & .34--.78 & \textbf{.48} [.437, .529] \\
\textsc{xCOMET-XXL}   & .32 & .28          & ---      & .26 [.223, .301] \\
Mistral-Large         & .17 & .23          & .19--.32 & .28 [.235, .317] \\
GPT-4o-mini           & .09 & .15          & .08--.25 & .21 [.163, .261] \\
LLaMA-4               & .08 & .13          & .03--.23 & .14 [.104, .183] \\
\bottomrule
\end{tabular}
\caption{Span-level agreement with the professional human MQM reference on the EU20 subset. OC columns use greedy one-to-one matching on target character offsets (OC $\geq 0.8$), averaged over 9 languages (25 items each). \emph{$F_1$ range} gives the per-language min--max Span-$F_1$. Char-$F_1$ is a threshold-free character-level span-overlap metric on the binary target-side error mask, reported as global micro with 95\% bootstrap CIs (per-language ranges are unavailable for \textsc{xCOMET-XXL}).}
\label{tab:agreement-main}
\end{table}

\paragraph{Span-level Agreement.}
Table~\ref{tab:agreement-main} shows that GPT-5.2 yields the highest agreement with the human reference under OC-based span matching (mean OC Span-F1 $=.55$), outperforming the other automatic annotators.
Among automatic baselines, xCOMET-XXL is stronger (mean OC Span-F1 $=.28$) than GPT-4o-mini and LLaMA-4 and broadly comparable to or slightly above Mistral-Large under our OC-based matching, but remains below GPT-5.2 on naturally occurring benchmark translations.
Additional OC-based pairwise comparisons using GPT-5.2 and xCOMET-XXL as reference annotators are reported in Appendix~\ref{sec:app:tqann:matching} (Table~\ref{tab:agreement-oc-extended}).
Appendix~\ref{sec:app:tqann:metric_examples} provides worked examples for OC and SIM, while Appendix~\ref{sec:app:tqann:examples} collects additional qualitative source-side and target-side example cases.
Appendix~\ref{sec:app:tqann:charoverlap} (Table~\ref{tab:annotator-stats}) further shows that xCOMET-XXL is more fine-grained than the other automatic annotators, producing more and shorter spans on average.

Besides GPT-5.2, Mistral-Large generally achieves higher agreement scores than GPT-4o-mini and LLaMA-4, with LLaMA-4 typically lowest.
This qualitative ranking is robust across alternative evaluation views: threshold-free character-overlap metrics and pooled bootstrap confidence intervals support the same overall conclusion, with GPT-5.2 remaining strongest and the remaining LLM annotators ordered similarly (Appendix~\ref{sec:app:tqann:charoverlap}, Table~\ref{tab:charoverlap-main}; Appendix~\ref{sec:app:tqann:bootstrap}, Table~\ref{tab:bootstrap-ci-llm}).
Appendix~\ref{sec:app:tqann:matching} further reports SIM-based agreement (Table~\ref{tab:agreement-sim}) and per-language Span-F1 for OC and SIM (Table~\ref{tab:f1-by-lang}), showing that relative annotator rankings are stable even though absolute agreement varies across language batches.
Appendix~\ref{sec:app:tqann:bootstrap} (Table~\ref{tab:bootstrap-ci-llm}) further reports pooled bootstrap confidence intervals over items, confirming the same qualitative ranking for the four LLM annotators.

The low agreement between the automatic annotators and the human reference indicates that span boundaries are often chosen differently and that annotators may also disagree on what should count as a translation error in the target text.

% Table: Source-overlap
\begin{table}[thb]
\centering
\tabcolsep=0.98em
\small
\begin{tabular}{lcc}
\toprule
\textbf{Annotator} & \textbf{Mean overlap rate} & \textbf{Range} \\
\midrule
GPT-5.2       & .06 & .00--.13 \\
Human         & .08 & .04--.13 \\
xCOMET-XXL    & .08 & .04--.13 \\
GPT-4o-mini        & .10 & .00--.25 \\
LLaMA-4       & .13 & .06--.29 \\
Mistral-Large       & .14 & .09--.20 \\
\bottomrule
\end{tabular}
\caption{Source-overlap: fraction of an annotator's target spans with valid offsets that overlap target regions linked to GPT-5.2 source-anomaly annotations. Values are means over 9 languages; ranges denote min--max across languages. Higher values indicate a stronger tendency to mark source-linked target regions as target-side errors (OC $\geq 0.8$; each annotator span counted at most once via its best overlap).}

\label{tab:source-overlap}
\end{table}

\paragraph{Source-overlap indicator.}
Table~\ref{tab:source-overlap} provides a diagnostic view of source-linked confounds.
GPT-5.2 has the lowest source-overlap rate, consistent with a stricter target-only annotation tendency.
xCOMET-XXL is close to the human rate, while Mistral-Large and LLaMA-4 show higher overlap, suggesting that they more often mark source-linked regions as target-side translation errors.

Human annotations are non-zero, reflecting that some source issues are preserved in the translation.
Per-language source-overlap rates are reported in Appendix~\ref{sec:app:tqann:matching} (Table~\ref{tab:source-overlap-by-lang}) and help contextualize language-dependent differences in span agreement.
More broadly, source-overlap helps explain part of the agreement differences across annotators: systems that frequently mark regions tied to source-side irregularities can diverge from annotators that focus more strictly on target-only errors.
Additional source-side and target-side example cases are collected in Appendix~\ref{sec:app:tqann:examples}.

\paragraph{Qualitative inspection.}
To contextualize the agreement scores, we manually spot-checked a small set of items across DE/DA/IT/RO.
Although limited, the inspection revealed recurring divergence patterns consistent with the quantitative results. GPT-4o-mini, LLaMA-4, and Mistral-Large sometimes label answer-option issues as \textit{Mistranslation} even when the translation is plausible, especially in multiple-choice settings.
Such cases are less frequent in the human reference and GPT-5.2.
Both human and LLM annotators also occasionally treat source-side ill-formedness as target-side translation errors, consistent with non-zero source-overlap rates (Appendix~\ref{sec:app:tqann:matching}, Table~\ref{tab:source-overlap-by-lang}), while occasional data irregularities (e.g., source--target mismatches) can reduce span agreement independently of annotation quality.
We also observed model-specific tendencies: GPT-5.2 sometimes reverts acceptable human post-edits and misses subtle meaning differences, whereas GPT-4o-mini can produce broader spans.
These observations suggest that disagreement is driven not only by span boundary variation but also by differing judgments about what should count as a target-side translation error, especially in the presence of source anomalies.
These findings motivate treating source anomalies as a confounder and reporting agreement both with and without source-linked regions in future work.

\begin{table*}[tbhp]
\centering
\tabcolsep=0.37em
\footnotesize
\renewcommand{\arraystretch}{1.05}
\begin{tabular}{l cccc cccc cccc}
\toprule
 & \multicolumn{4}{c}{GPT-4o-mini} & \multicolumn{4}{c}{LLaMA-4} & \multicolumn{4}{c}{Mistral-Large} \\
\cmidrule(lr){2-5}\cmidrule(lr){6-9}\cmidrule(lr){10-13}
\textbf{Aggregation} &
\textbf{F1} & \textbf{R} & $\mathbf{F1_t}$ & $\mathbf{R_t}$ &
\textbf{F1} & \textbf{R} & $\mathbf{F1_t}$ & $\mathbf{R_t}$ &
\textbf{F1} & \textbf{R} & $\mathbf{F1_t}$ & $\mathbf{R_t}$ \\
\midrule
\multicolumn{13}{@{}l}{\textit{Accuracy (20 EU20 target languages)}} \\[1pt]
\quad mean (N)   & .10/.21 & .17/.23 & .33/.48 & .57/.55 & .13/.10 & .17/.12 & .34/.31 & .43/.36 & .24/.34 & .36/.42 & .47/.60 & .70/.71 \\
\quad mean (cap) & .17/.28 & .25/.31 & .31/.46 & .49/.51 & .21/.20 & .26/.22 & .34/.35 & .43/.40 & .28/.39 & .39/.45 & .41/.57 & .59/.65 \\
\midrule
\multicolumn{13}{@{}l}{\textit{Fluency / Style (German-only, four anaphoric phenomena)}} \\[1pt]
\quad mean (N)   & .05/.32 & .09/.37 & .41/.57 & .68/.67 & .15/.16 & .19/.17 & .42/.41 & .59/.47 & .45/.46 & .62/.54 & .55/.56 & .82/.67 \\
\quad mean (cap) & .04/.32 & .08/.36 & .47/.61 & .74/.71 & .12/.14 & .16/.15 & .40/.41 & .58/.49 & .47/.48 & .62/.56 & .58/.59 & .83/.70 \\
\bottomrule
\end{tabular}
\caption{\RefSpanACES{} span localization across subsets. Classic matching (Span-F1 $F1$, Span-Recall $R$) requires exact span equality; tolerant matching ($F1_t$, $R_t$) allows gold-centered boundary slack up to $k{=}3$ tokens per side. Each cell reports \emph{Baseline/Ours} (initial GEMBA-ESA vs.\ our updated prompt).}
\label{tab:refspanaces-combined}
\end{table*}

\section{\RefSpanACES{}}
\label{sec:RefSpanACES}
To complement the EU20 study with an external gold-span benchmark, we evaluate the same LLM annotators on \RefSpanACES{}, a cleaned projection of Span-ACES.

\subsection{Methodology}
\label{sec:refspanaces:methodology}
\paragraph{\RefSpanACES{} construction.}
A key limitation of SPAN-ACES is that it labels only the introduced error span per item, leaving any additional translation errors unlabeled.
Consequently, span-based annotators that correctly flag such extra errors are counted as false positives, biasing precision and Span-F1 downward.
Each item consists of a human reference, a good translation, and an incorrect translation containing exactly one targeted phenomenon.
We construct \RefSpanACES{} by extracting the single contentful token-level difference between the good and incorrect translations, mapping that edit to the corresponding location in the human reference, and discarding items with multiple diffs or ambiguous matches.
This reduces unlabeled noise and yields more reliable span-level precision and Span-F1.

\paragraph{Mapping to MQM categories.}
Using the ACES authors' released mapping, we collapse phenomena into the two coarse MQM categories \textit{Accuracy} and \textit{Fluency/Style}, yielding 1{,}155 \textit{Accuracy} instances across the 20 EU20 target languages and 252 \textit{Fluency/Style} instances for German only in \RefSpanACES{}.
Dataset composition follows the final \RefSpanACES{} split described above, and the MQM mapping is reported in Appendix~\ref{sec:app:tqrefspanaces:details} (Table~\ref{tab:span-aces-ref-mapping}).

\paragraph{Dataset validation (human vs.~GPT-5.2).}
To validate the transformation pipeline and identify projection errors, we manually reviewed 178 \RefSpanACES{} records from a stratified sample using a fixed checklist and compared the resulting \texttt{pass}/\texttt{fail} judgments with GPT-5.2.
The two assessments agreed on 165/178 items (0.93), suggesting that most cases are straightforward under these criteria and that GPT-5.2 can provide a useful auxiliary signal for scalable dataset validation.
The remaining 13 disagreements were mostly borderline cases, such as type drift after projection or subtle side effects.

\paragraph{Span metrics (classic vs.~tolerant).}
We report classic span metrics (Span-F1 $F1$ and Span-Recall $R$) and a tolerant variant ($F1_t$, $R_t$) that allows small boundary mismatches without overlap thresholds. Classic matching requires exact equality between predicted and gold spans.
Tolerant matching is gold-centered: a prediction is counted as correct if it contains the gold span and may extend by up to $k$ tokens on either side.
Metrics are first averaged within each phenomenon and then aggregated per MQM category using either sample-count weighting (\textbf{mean (N)}) or capped weighting (\textbf{mean (cap)}, $C{=}25$) to limit dominance by large phenomena.

\paragraph{Baseline vs.~our prompt.}
We evaluate two prompting setups on \RefSpanACES{}: (i) the initial GEMBA-ESA baseline prompt, and (ii) our updated MQM-style prompting (Section~\ref{sec:tqann:methodology}, paragraph ``LLM-based Annotation'').
Unless stated otherwise, tables report each metric as \emph{Baseline/Ours}.

\subsection{Results and Discussion}
\label{sec:refspanaces:results}

\paragraph{Accuracy and Fluency/Style.}
Table~\ref{tab:refspanaces-combined} reports span localization results for \textit{Accuracy} and \textit{Fluency/Style}.
On \textit{Accuracy}, Mistral-Large performs best overall, and tolerant matching substantially improves scores for all models, indicating frequent boundary near-misses.
Under mean (N), the updated prompt raises classic F1 for GPT-4o-mini (\textbf{.10$\to$.21}) and Mistral-Large (\textbf{.24$\to$.34}), but slightly lowers it for LLaMA-4 (\textbf{.13$\to$.10}), a pattern that also holds under tolerant matching and therefore suggests prompt--model sensitivity.
On \textit{Fluency/Style}, scores are higher overall, but because this subset is German-only and limited in scope, we view it mainly as a controlled diagnostic.
Here too, prompt refinements help GPT-4o-mini most strongly (classic F1 \textbf{.05$\to$.32}, mean (N)), while gains for Mistral-Large (\textbf{.45$\to$.46}) and LLaMA-4 (\textbf{.15$\to$.16}) are marginal, indicating that prompt improvements do not transfer uniformly across architectures.
Across models and categories, tolerant scores ($F1_t$, $R_t$) are consistently higher than classic scores, indicating frequent boundary near-misses rather than complete detection failures.
This matches a common pattern in LLM annotations: different systems often point to the same error region but choose slightly different span boundaries (e.g., broader or shifted spans).

\section{Performance Impact Analysis}
\label{sec:tqperf}
To assess how translation quality relates to downstream LLM performance, we link span-level error annotations to whether models answer translated instances correctly and analyze these associations with logistic regression.

\subsection{Methodology}
\label{sec:tqperf:methodology}

\paragraph{Data and annotations.}
For our performance analysis, we use MMLU, ARC, and GSM8K from the EU20 benchmark suite and annotate target-side translation errors with an MQM-inspired prompting scheme (Section~\ref{sec:tqann}) using four LLM-based annotators: GPT-5.2, GPT-4o-mini, LLaMA-4, and Mistral-Large.
For robustness, we additionally include xCOMET-XXL as non-generative baseline in the same regression pipeline.
To control for issues already present in the original English items (e.g., inconsistencies or ambiguities that may affect model accuracy independent of translation), we additionally annotate the English source for a paired subset of 651 translated instances (item--language pairs) using GPT-5.2.
Together with the corresponding target-side annotations, this paired subset enables a direct comparison of source-side issues ($S$) versus translation-induced errors ($T$).
For the performance-impact regressions, we exclude HellaSwag and TruthfulQA: in HellaSwag, many ``odd'' endings are intentionally constructed distractors (task design rather than source noise), which would blur the interpretation of $S$, and for TruthfulQA we could not reliably derive binary correctness labels from the released evaluation logs.
The resulting paired subset contains 651 item--language pairs and 4{,}993 translated observations across the evaluated models. Detailed dataset statistics, including translated and English correctness counts and annotation sparsity, are reported in Appendix~\ref{sec:app:tqperf:data} (Table~\ref{tab:tqperf_data_stats}).

\paragraph{Model evaluation.}
We evaluate eight instruction-tuned multilingual LLMs (Appendix~\ref{sec:app:tqperf:llms}, Table~\ref{tab:models}) using a modified version of EleutherAI's LM Evaluation Harness\footnote{\href{https://github.com/EleutherAI/lm-evaluation-harness}{github.com/EleutherAI/lm-evaluation-harness}} and record a binary outcome (correct/incorrect) for each item in English and in each translated variant.

\paragraph{Analysis dataset.}
We merge per-instance correctness with the corresponding span-level error annotations to construct an analysis dataset that links (i) the presence of source-side issues and/or translation errors to (ii) each evaluated LLM's correctness on the English original and its translated variants. We analyze this dataset with logistic regression, focusing on the paired-annotation subset when comparing translation errors against source-side issues.

\paragraph{Regression setup.}
We use logistic regression to estimate how the presence of translation errors ($T$) and source-side issues ($S$) predicts correctness on translated items, while controlling for whether the model solves the English original and for systematic differences across languages, datasets, and evaluation models.
A key property of our data is that for each underlying item and evaluation model we observe both (i) performance on the English original and (ii) performance on its EU20 translations, enabling a clean separation between English-level ability and translation-associated failures.

\paragraph{Instances and variables.}
Our unit of analysis is a translated instance: we record whether evaluation model $m$ answers an item in target language $\ell$ correctly ($y\in\{0,1\}$; English excluded).
For the same item and model, we record whether the English original is answered correctly, denoted $y^{EN}\in\{0,1\}$.
From the span-level annotations, we derive two binary indicators:
$T=1$ if the translation contains at least one annotated target-side error (annotator-specific), and $S=1$ if at least one issue was annotated in the English source.
%$T=1$ if the translation contains at least one annotated target-side error (for the respective target-side annotator), and $S=1$ if at least one issue was annotated in the English source (GPT-5.2).
Since $S$ refers to the English original, it is shared across all translations of the same item.

\paragraph{Specifications.}
We fit logistic regressions with fixed effects for target language, dataset, and evaluation model, denoted by $C$.

\textbf{(A) Full model with English control.}
This specification asks: \emph{How strongly are translation errors associated with lower accuracy after controlling for whether the model solves the English original?}
\begin{equation}
\Pr(y=1)=\sigma\!\big(\beta_T T+\beta_S S+\beta_{EN} y^{EN}+C\big).
\end{equation}

\textbf{(B) Only items solved in English.}
Estimated on the subset with $y^{EN}=1$, this specification asks:
\emph{Among items the model answers correctly in English, how are translation errors ($T$) and source issues ($S$) associated with accuracy in translation?}
\begin{equation}
\Pr(y=1)=\sigma\!\big(\beta_T T+\beta_S S+C\big).
\end{equation}

\textbf{Ablations (omit $S$).}
To assess sensitivity to omitted-variable bias, we refit (A) and (B) without $S$. If translation errors correlate with source-side issues that also depress accuracy, omitting $S$ can make the estimated translation-error association more negative.

\paragraph{Effect reporting and uncertainty.}
For each target annotator and specification, we report average marginal effects (AMEs) of $T$ and $S$ in probability points (more interpretable than log-odds).
We compute confidence intervals using a block bootstrap over underlying items: we resample items with replacement, include all associated rows (across target languages and evaluation models) for each resampled item, refit the model, and take percentile intervals from the bootstrap distribution.
We also run a counterfactual ranking analysis by predicting each evaluation model's translated accuracy under $T{=}0$ and comparing the resulting ranking to the observed ranking (Appendix~\ref{sec:app:tqperf:ranking}).

\paragraph{Counterfactual ranking.}
To assess whether translation errors affect not only absolute scores but also relative system ordering, we additionally predict each evaluation model's translated accuracy under the counterfactual setting $T{=}0$ and compare the resulting ranking to the observed ranking.

\subsection{Results and Discussion}
\label{sec:tqperf:results}

\begin{figure*}[t]
  \centering
  \includegraphics[width=\textwidth]{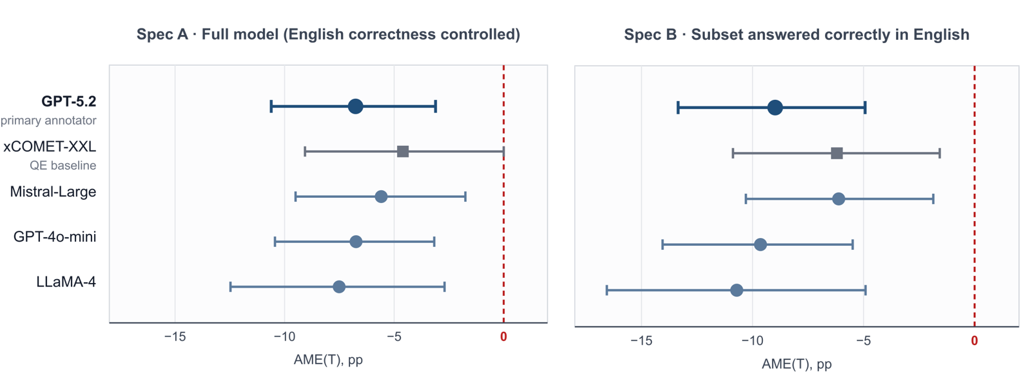}
  \caption{Average marginal effects (AMEs) of target-side translation errors $T$ on translated correctness. Panel~A shows the full model controlling for English correctness; Panel~B is restricted to items answered correctly in English ($y^{\mathrm{EN}}{=}1$). Points show AME($T$) in probability points (pp), bars 95\% block-bootstrap confidence intervals over underlying items, and the dashed vertical line marks zero. Across annotators, translation errors are associated with lower translated accuracy; \textsc{xCOMET-XXL} shows the same negative direction and is clearly below zero in Spec.~B.}
  \label{fig:tqperf_ame}
\end{figure*}

Figure~\ref{fig:tqperf_ame} summarizes the estimated effect of target-side translation errors on translated accuracy under the two main specifications.
xCOMET-XXL yields the same qualitative pattern as baseline annotator: its estimated translation-error effect is also negative, borderline in the full model (Spec.~A: AME($T$) = \textbf{-4.60} pp, 95\% CI \textbf{[-9.07, 0.00]}) but clearly below zero in the English-correct subset (Spec.~B: \textbf{-6.20} pp, 95\% CI \textbf{[-10.88, -1.57]}).
This indicates that the main result does not depend on a single LLM-as-a-judge annotation approach.
Across all four annotators, target-side translation errors ($T$) are consistently associated with lower correctness in translation, even when controlling for source-side issues ($S$), English correctness, and fixed effects.
In the full model with an English control (Spec.~A), AME($T$) ranges from \textbf{-5.59 to -7.51} pp and all 95\% bootstrap CIs lie entirely below zero.
Conditioning on items solved in English (Spec.~B; $y^{EN}{=}1$) yields larger drops, with AME($T$) between \textbf{-6.12 and -10.71} pp, again with CIs fully below zero.
Together, A and B provide complementary evidence that translation errors are associated with accuracy losses and that these losses persist even when the underlying item is solvable in English.

Source-side issues ($S$) are directionally negative but weaker and less stable.
In Spec.~A, AME($S$) is around \textbf{-3.5} to \textbf{-3.8} pp, but the corresponding CIs generally overlap zero, consistent with $y^{EN}$ absorbing much of the item-difficulty signal.
In Spec.~B, AME($S$) is somewhat larger and becomes significant for some annotators, suggesting that source anomalies can additionally depress translated performance once we restrict to English-solvable items.
Omitting $S$ changes AME($T$) only marginally (typically about \textbf{0.2--0.5} pp), so the translation-error effect is not simply explained away by source-side issues.
Logit coefficients and odds ratios mirror these patterns (Appendix Tables~\ref{tab:tqperf_logit_coef} and~\ref{tab:tqperf_logit_or}), with OR($T$) consistently below 1 (roughly .65--.72 in A and .54--.69 in B).
Full AME estimates for all specifications, including xCOMET-XXL, are reported in Appendix~\ref{sec:app:tqperf:logreg} (Tables~\ref{tab:tqperf_ame_specA} and~\ref{tab:tqperf_ame_specB}).

% Message: Übersetzungsfehler kosten pro betroffenem Item ~6--11 pp. Bei den beobachteten Fehlerhäufigkeiten entspricht das grob ~2--6 pp auf Dataset-Ebene (je nach Annotator und Spezifikation).
\paragraph{Practical impact.}
Figure~\ref{fig:tqperf_ame} reports effects in probability points (pp), which can be read directly as accuracy drops. Across annotators, a translation with at least one annotated error ($T{=}1$) is associated with a \textbf{6--8 pp} lower probability of a correct answer in the full model (Spec.~A) and a \textbf{6--11 pp} lower probability when restricted to items solved in English (Spec.~B; $y^{EN}{=}1$).
Thus, translation errors are associated with a noticeable loss in measured accuracy even when the underlying item is solvable in English.

To translate these per-item drops into overall impact, we use the approximation \emph{overall loss} $\approx \Pr(T{=}1)\times|\mathrm{AME}(T)|$.
With GPT-5.2 annotations, $T{=}1$ occurs in \textbf{63\%} of translated observations in Spec.~A; combined with AME($T$)$=\textbf{-6.76}$ pp, this implies an overall loss of about \textbf{4.3 pp}.
In the English-correct subset of Spec.~B, the same calculation gives larger impacts (e.g., GPT-5.2: \textbf{62\%}$\times$\textbf{8.98} pp $\approx$ \textbf{5.6 pp}).
These figures are intended as a readable approximation of the observed association, not a separate causal estimand.

For intuition, every additional \textbf{10\%} of items with a translation error corresponds to about $|\text{AME}(T)|/10$ points of accuracy loss: roughly \textbf{0.6--0.8 pp} in Spec.~A and \textbf{0.6--1.1 pp} in Spec.~B.
This approximation ignores overlap between $T$ and source-side issues $S$ beyond what is already captured by the regression controls.

\paragraph{Counterfactual ranking.}
To assess whether translation errors affect not only absolute scores but also relative system ordering,
we compare the observed ranking of evaluation models to a counterfactual ranking obtained by predicting translated accuracy under $T{=}0$.
Across the three annotators included in this auxiliary analysis (Mistral-Large, LLaMA-4, and GPT-5.2), the observed and counterfactual rankings are nearly identical: Spearman's $\rho$ is essentially 1.0 in all cases, with equally tight Kendall $\tau$ estimates (Appendix~\ref{sec:app:tqperf:ranking}, Table~\ref{tab:tqperf_rankcorr}).
This suggests that translation errors behave like an approximately uniform penalty across systems: they shift absolute scores downward, but usually do not induce rank swaps.
At the same time, the corresponding counterfactual score uplifts remain non-trivial---roughly 1--4.6 percentage points depending on annotator and evaluation model (Appendix~\ref{sec:app:tqperf:ranking}, Table~\ref{tab:tqperf_rankuplift})---so ranking stability does not remove the broader problem of biased absolute measurements.

\section{Conclusion}
We study how translation errors in machine-translated benchmarks affect multilingual LLM evaluation by combining span-level MQM-style error annotation with downstream performance modeling.
On an EU20 subset with a professional human reference (225 items, nine target languages), GPT-5.2 shows the highest agreement with humans, while xCOMET-XXL provides the strongest non-generative baseline.
This ranking is supported not only by thresholded span matching, but also by threshold-free character-overlap and unitizing-reliability analyses, indicating that the main agreement result is not a metric artifact.

We further evaluate span localization on \RefSpanACES{} (1{,}407 items), where tolerant span metrics are substantially higher than classic ones, suggesting that many residual disagreements are boundary near-misses rather than complete detection failures.
Finally, in regressions with English controls and explicit source-side issue indicators, target-side translation errors are consistently associated with lower translated accuracy (about 6--8 pp in the full model and 6--11 pp when restricting to items solved in English), while source-side issues have smaller and less stable effects.
A counterfactual ranking analysis further shows that translation errors mainly shift absolute scores rather than relative system rankings: leaderboard orderings are usually preserved, but absolute performance estimates remain biased downward.
Our results suggest that translated leaderboards can remain internally stable while still understating absolute multilingual performance.
Future work should extend this analysis to finer-grained error types and multi-span phenomena, and develop more boundary-robust span evaluation and validation protocols for translation-aware benchmark release.

\section*{Limitations}
\label{sec:lim}
First, high-quality span-annotated references remain scarce, particularly for lower-resource languages and for fine-grained MQM error inventories. Our EU20 human reference covers nine target languages and 225 items, with one professional annotator per language, so human-human inter-annotator agreement is not available. To partially address this, we report a shared-subset reliability analysis on the 225-segment EU20 subset using the available automatic annotators plus the human reference; however, we do not present this as a replacement for human-human IAA. \RefSpanACES{} provides an additional external benchmark, but it covers controlled phenomena rather than naturally occurring error distributions.

Second, span-level MQM annotation has no widely accepted cross-lingual gold standard, and expert annotation is costly; while we validate parts of our setup (human reference checks and a \RefSpanACES{} validation subset), residual subjectivity in span boundaries and in what counts as a target-side error is unavoidable.

Third, our performance-impact analysis is correlational: although we control for English correctness, fixed effects, and source-side issues, unobserved confounders and model/specification choices can still affect effect sizes, and some signals (especially source-side issues) are sparse and therefore estimated with greater uncertainty.

Fourth, automatic annotation quality depends on the chosen models and prompting setup. 
We mitigate this by comparing multiple LLM annotators, including another span-aware baseline (xCOMET-XXL), and by reporting robustness analyses across alternative agreement views; however, conclusions about the strongest annotator may still not transfer to other model families, prompting strategies, or annotation interfaces.

Finally, our results reflect a specific selection of benchmarks (EU20 tasks), MT system, source and target languages, and evaluated LLMs; translation artifacts and their downstream impact may differ in other languages, domains, genres, or translation pipelines.

\section*{Acknowledgments}
\label{sec:ack}
This work was funded by the German Federal Ministry for Economic Affairs and Climate Action (BMWK) through the OpenGPT-X project (no.~68GX21007D). The authors further acknowledge support from the Federal Ministry of Research, Technology and Space of Germany (BMFTR), the Sächsische Staatsministerium für Wissenschaft, Kultur und Tourismus through ScaDS.AI (Center for Scalable Data Analytics and Artificial Intelligence Dresden/Leipzig), and the BMFTR under grant no.~01IS24077A. We also thank the Gauss Centre for Supercomputing e.V. for computing time on JUWELS at Jülich Supercomputing Centre (JSC), and ZIH at TU Dresden for infrastructure used in automatic evaluation computations. GNU parallel \citep{tange_2024} supported parts of the data processing.

\bibliography{literature}

\appendix

\input{appendix/tqann}
\input{appendix/tqrefspanaces}
\input{appendix/tqperf}

\end{document}

%% file: appendix/tqann.tex
\section{MQM Annotation}
\label{sec:app:tqann}

\subsection{Annotation Guide and Interface}
\label{sec:app:tqann:guide}

This subsection summarizes the manual protocol used to create our human reference and documents the annotation interface.
Our guidelines are based on the Multidimensional Quality Metrics (MQM) framework,\footnote{\url{https://themqm.org/the-mqm-full-typology/}} using a reduced, task-focused typology.
Concretely, we group error labels into two high-level dimensions---\textit{Accuracy} (semantic faithfulness) and \textit{Fluency/Style} (well-formedness and naturalness)---and annotate errors as spans.
For each error, annotators highlight the erroneous span in the target translation and, where applicable, the corresponding source span to support span-level alignment analyses.
We treat \textit{Addition} and \textit{Omission} as asymmetric cases: \textit{Addition} is annotated only on the target side (no source span), whereas \textit{Omission} is annotated only on the source side (no target span).
Table~\ref{tab:subset-stats} summarizes the sampled EU20 subset, and Tables~\ref{tab:mqm-labels-accuracy} and~\ref{tab:mqm-labels-fluency} list the reduced MQM inventory used in the manual annotation.

\begin{table}[tbhp]
\centering
\footnotesize
\setlength{\tabcolsep}{3pt}
\begin{tabular}{lrrrrr}
\toprule
\textbf{Dataset} & \textbf{k/lang} & \textbf{Ex.} & \textbf{Tok.} & \textbf{Tok. min--max} & \textbf{Sent.} \\
\midrule
ARC        & 4 & 36 & 56.6  & 31--98   & 5.0 \\
GSM8K      & 2 & 18 & 129.5 & 93--191  & 8.3 \\
HellaSwag  & 9 & 81 & 166.2 & 112--253 & 12.3 \\
MMLU       & 6 & 54 & 153.5 & 62--402  & 8.7 \\
TruthfulQA & 4 & 36 & 99.3  & 66--201  & 10.5 \\
\bottomrule
\end{tabular}
\caption{Aggregated statistics for the manually annotated EU20 subset across nine languages. \textbf{k/lang} is the sampling quota per language, yielding \textbf{Ex.} total instances per dataset across all languages. \textbf{Tok.} and \textbf{Sent.} report the mean number of tokens and sentences per instance, respectively. \textbf{Tok. min--max} gives the observed token-length range within each dataset.}
\label{tab:subset-stats}
\end{table}

\begin{table}[tbhp]
\centering
\tabcolsep=0.28em
\footnotesize
\begin{tabular}{ll}
\toprule
\textbf{Label} & \textbf{Description} \\
\midrule
Addition & Adds content not supported by source \\
Omission & Drops source content \\
Mistranslation & Meaning is changed or incorrect \\
Under-translation & Meaning is too vague; nuance is lost \\
Over-translation & Adds unwarranted specificity or detail \\
Reordering & Changes attachment or meaning \\
Untranslated & Leaves a source fragment untranslated \\
Wrong language & Uses tokens from unintended language \\
Do-not-translate & Content that should remain unchanged \\
\bottomrule
\end{tabular}
\caption{Reduced MQM label set for \textit{Accuracy} errors used in manual annotation.}
\label{tab:mqm-labels-accuracy}
\end{table}

\begin{table}[tbhp]
\centering
\tabcolsep=0.25em
\footnotesize
\begin{tabular}{ll}
\toprule
\textbf{Label} & \textbf{Description} \\
\midrule
Grammar & Grammatical error (agreement, tense, case) \\
Spelling & Spelling or character error \\
Punctuation & Punctuation error affecting readability \\
Inconsistent & Inconsistent terminology or naming \\
Awkward & Unnatural or non-idiomatic phrasing \\
Unintelligible & Output is not reliably interpretable \\
\bottomrule
\end{tabular}
\caption{Reduced MQM label set for \textit{Fluency/Style} errors used in manual annotation.}
\label{tab:mqm-labels-fluency}
\end{table}

Annotators assign one of two severity levels to each annotated span. \textit{Major} errors change meaning, can mislead, or substantially harm understanding/faithfulness, whereas \textit{Minor} errors largely preserve meaning and primarily affect fluency/style or constitute a limited local defect. If no label fits or annotators are uncertain, they mark the span as \textit{Other/Unknown} (used sparingly).

\begin{table}[tbhp]
\centering
\footnotesize
\setlength{\tabcolsep}{4pt}
\begin{tabularx}{\columnwidth}{p{9em} p{7em} X}
\toprule
\textbf{System} & \textbf{Type} & \textbf{Reference} \\
\midrule
GPT-5.2 & LLM & \href{https://platform.openai.com/docs/models/gpt-5.2}{OpenAI model page} \\
GPT-4o-mini & LLM & \href{https://platform.openai.com/docs/models/gpt-4o-mini}{OpenAI model page} \\
LLaMA-4 Scout 17B & LLM & \href{https://huggingface.co/meta-llama/Llama-4-Scout-17B-16E-Instruct}{Hugging Face model card} \\
Mistral-Large-Instruct-2411 & LLM & \href{https://huggingface.co/mistralai/Mistral-Large-Instruct-2411}{Hugging Face model card} \\
\textsc{xCOMET}-XXL & QE / span baseline & Guerreiro et al.~(2024b) \\
\bottomrule
\end{tabularx}
\caption{Automatic annotators used in the appendix analyses. The four LLMs are evaluated throughout the appendix; \textsc{xCOMET}-XXL is additionally included wherever the auxiliary analysis covers the non-generative span baseline.}
\label{tab:annotator-models}
\end{table}

\begin{figure*}[t]
    \centering
    \includegraphics[width=0.98\textwidth]{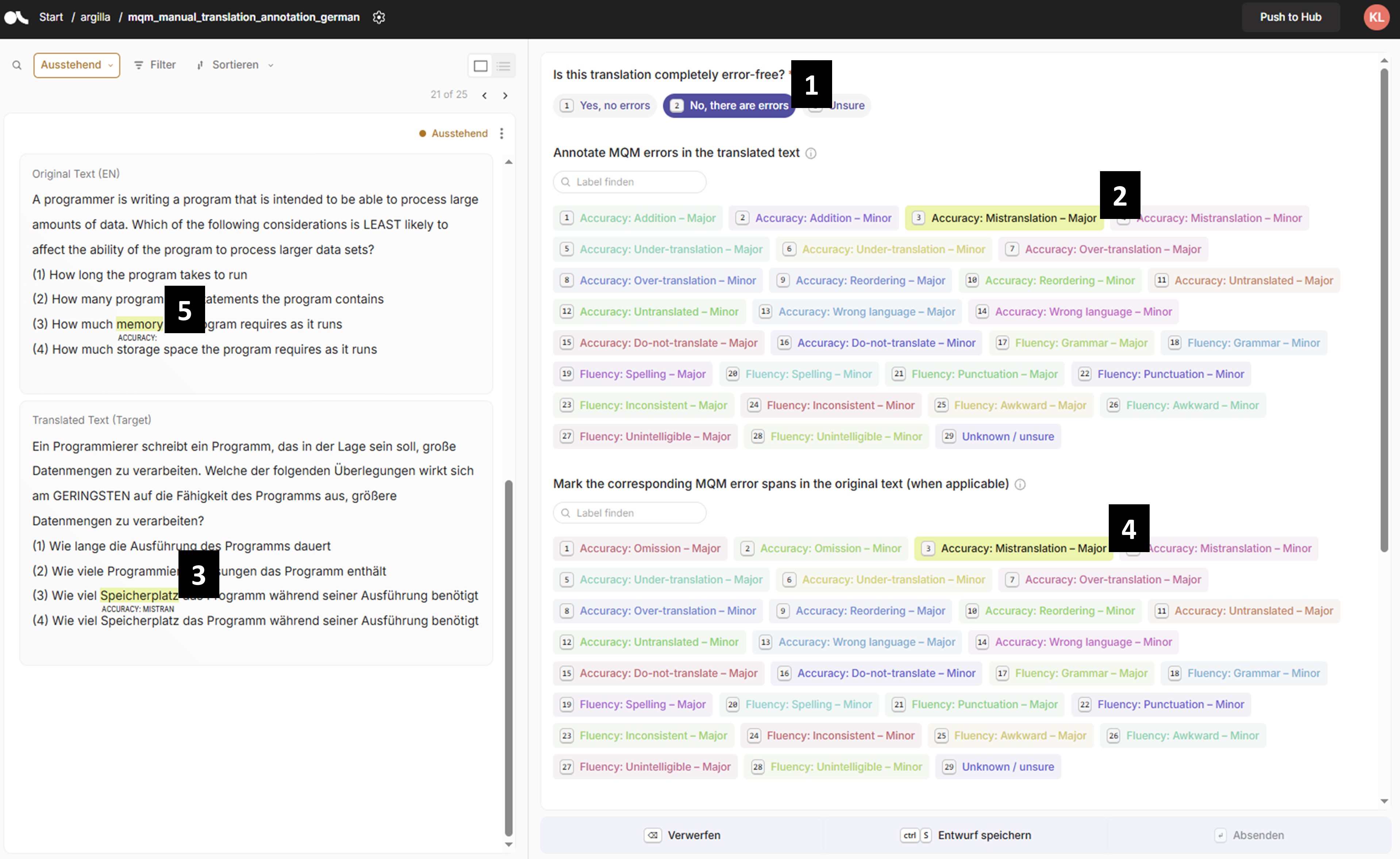}
    \caption{Argilla interface for span-based MQM annotation. The numbered markers indicate the main interaction elements: (1) error-free decision, (2) target-side MQM label selection, (3) target-span highlighting, (4) source-span highlighting, and (5) the source/context panel.}
    \label{fig:argilla-ui-annot}
\end{figure*}

\begin{figure*}[t]
    \centering
    \includegraphics[width=0.98\textwidth]{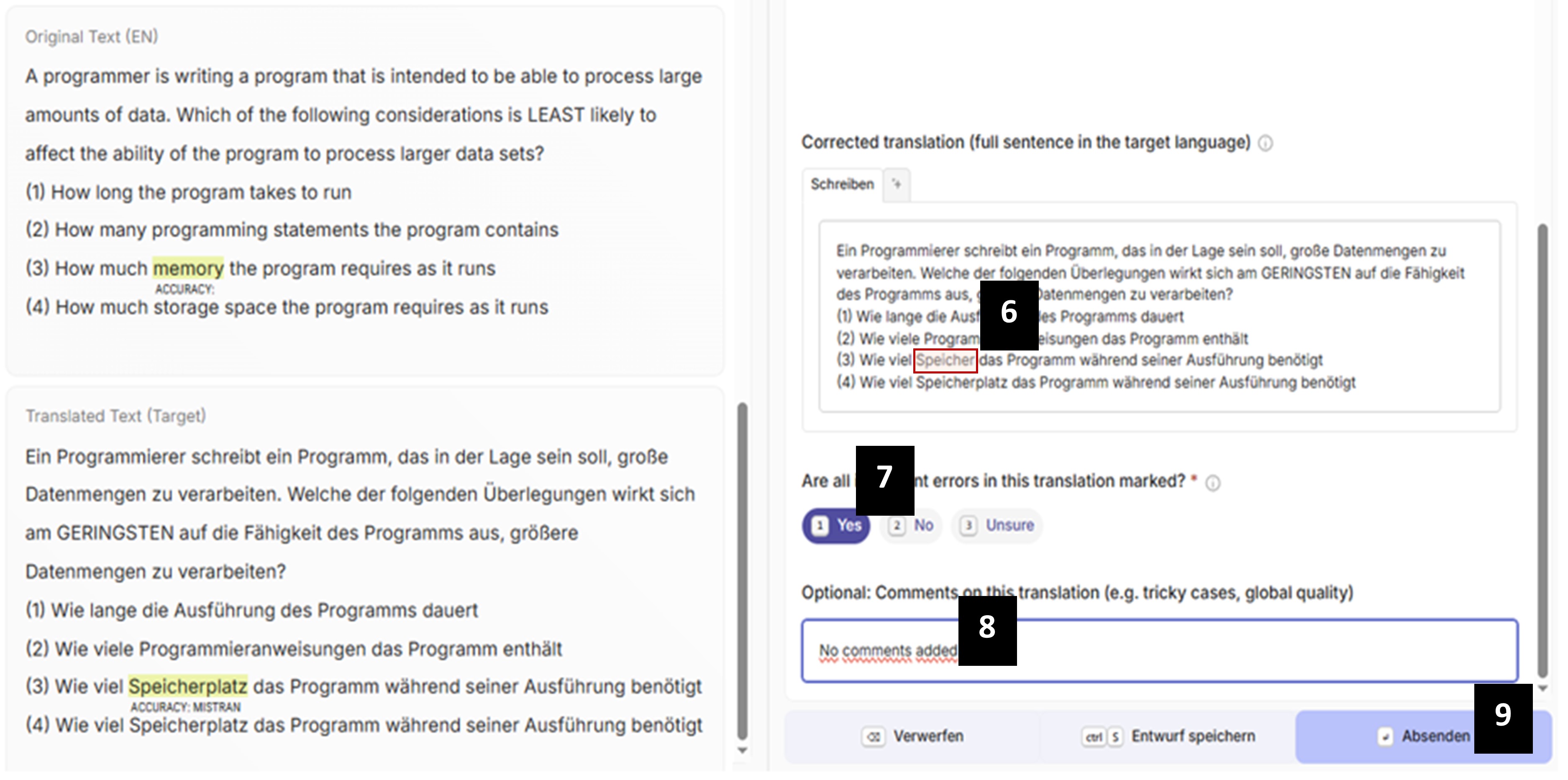}
    \caption{Argilla interface for post-editing and metadata. The numbered markers indicate (1) the minimally corrected translation field, (2) the completeness control question, (3) optional annotator comments, and (4) final submission.}
    \label{fig:argilla-ui-postedit}
\end{figure*}

Figure~\ref{fig:argilla-ui-annot} and Figure~\ref{fig:argilla-ui-postedit} illustrate the Argilla-based interface and the annotation workflow. Concretely, annotators proceeded as follows:

\begin{enumerate}
    \item Decide whether the translation is error-free (\textit{Yes/No/Unsure}).
    \item If errors are present, mark erroneous target spans and assign an MQM label and severity.
    \item Mark corresponding source spans where applicable; \textit{Addition} is annotated only on the target side, whereas \textit{Omission} is annotated only on the source side.
    \item Provide a minimally post-edited corrected translation.
    \item Answer a control question indicating whether all important errors were captured.
    \item Optionally add comments for clarification.
\end{enumerate}

\subsection{Metrics and Matching}
\label{sec:app:tqann:matching}

We compare span sets using (i) a position-based overlap coefficient (OC) computed from target character offsets, and (ii) a string-based similarity metric (SIM) defined as raw character 3-gram Dice similarity, without text normalization. OC and SIM capture complementary notions of span agreement: OC is offset-based and rewards positional overlap, whereas SIM is text-based and remains applicable when offsets are missing or boundaries drift. Offsets are validated against the raw target string. If missing or inconsistent, we attempt to recover them via exact substring search and leave offsets missing in ambiguous cases.
Not all auxiliary analyses in this appendix cover the same system set: where relevant, we report xCOMET-XXL alongside the LLM annotators, but some robustness analyses are restricted to the four LLM-based comparisons.

For character-offset spans $a$ and $b$, we define the overlap coefficient as
\[
\mathrm{OC}(a,b) = \frac{|a \cap b|}{\min(|a|, |b|)},
\]
i.e., the character-overlap length divided by the shorter span length.
For OC-based matching, a gold span and a predicted span are matchable if $\mathrm{OC} \ge \tau_{\mathrm{OC}}$; in the main text we use $\tau_{\mathrm{OC}}=0.8$.

For span texts $s_1$ and $s_2$, let $\mathcal{G}_3(s)$ denote the multiset of raw character trigrams of $s$. We define
\[
\mathrm{SIM}(s_1,s_2)=\frac{2\sum_{g}\min(c_g(s_1),c_g(s_2))}
{\sum_{g} c_g(s_1)+\sum_{g} c_g(s_2)},
\]
i.e., Sørensen--Dice similarity over character trigrams, where $c_g(s)$ is the count of trigram $g$ in $s$. For SIM-based matching, a gold span and a predicted span are matchable if $\mathrm{SIM} \ge \tau_{\mathrm{SIM}}$; in the main text we use $\tau_{\mathrm{SIM}}=0.6$.

OC matching requires valid target offsets. Spans without offsets cannot be matched and contribute to FP/FN. For SIM-based matching, we apply the metric to spans deduplicated by exact target-span text on both sides.

To compute precision/recall/F1, we perform greedy one-to-one matching: all matchable span pairs are sorted by score (descending) and selected if neither span has been matched before. For each comparison, we aggregate TP/FP/FN across samples to obtain micro-precision, micro-recall, and micro-F1. For sensitivity analysis, we additionally sweep thresholds over OC $\in \{0.7,0.8,0.9\}$ and SIM $\in \{0.4,0.5,0.6\}$.

\begin{table}[htpb]
\centering
\small
\setlength{\tabcolsep}{6pt}
\begin{tabular}{lcc}
\toprule
\textbf{Comparison} & \textbf{SIM Recall} & \textbf{SIM F1} \\
\midrule
Human vs GPT-5.2          & .38 & .45 (.30--.78) \\
Human vs Mistral-Large          & .14 & .17 (.10--.26) \\
Human vs GPT-4o-mini      & .11 & .15 (.09--.25) \\
Human vs LLaMA-4          & .06 & .10 (.02--.19) \\
GPT-5.2 vs Mistral-Large        & .16 & .17 (.09--.24) \\
GPT-5.2 vs GPT-4o-mini    & .15 & .21 (.11--.30) \\
GPT-5.2 vs LLaMA-4         & .09 & .12 (.03--.19) \\
\bottomrule
\end{tabular}
\caption{Span-level agreement based on string similarity (SIM) averaged over nine languages (25 items each). 
We report mean Span-Recall and Span-F1 across languages. 
Parentheses denote the min--max Span-F1 over languages. 
SIM uses raw character 3-gram Dice with threshold $0.6$ and greedy one-to-one matching on spans deduplicated by exact target-span text.
}
\label{tab:agreement-sim}
\end{table}

To complement the human-reference comparison in the main text, Table~\ref{tab:agreement-oc-extended} reports additional OC-based pairwise span agreement using GPT-5.2 and xCOMET-XXL as reference annotators. Because Span-F1 is symmetric but recall is directional, these auxiliary views help contextualize agreement patterns beyond the human-reference setting.

The per-language values in Table~\ref{tab:f1-by-lang} are descriptive and should be interpreted cautiously, given the small per-language sample size (25 items per language).

\begin{table}[htpb]
\centering
\tabcolsep=0.26em
\footnotesize
\begin{tabular}{lcc}
\toprule
\multicolumn{3}{c}{\textbf{(A) Human as reference}} \\
\midrule
Comparison & OC Recall & OC Span-F1 \\
\midrule
Human vs GPT-5.2       & .44 & .55 \\
Human vs xCOMET-XXL    & .32 & .28 \\
Human vs Mistral-Large & .17 & .23 \\
Human vs GPT-4o-mini   & .09 & .15 \\
Human vs LLaMA-4       & .08 & .13 \\
\midrule
\multicolumn{3}{c}{\textbf{(B) GPT-5.2 as reference}} \\
\midrule
Comparison & OC Recall & OC Span-F1 \\
\midrule
GPT-5.2 vs xCOMET-XXL  & .42 & .26 \\
GPT-5.2 vs Mistral-Large & .24 & .25 \\
GPT-5.2 vs GPT-4o-mini & .15 & .21 \\
GPT-5.2 vs LLaMA-4     & .12 & .17 \\
\midrule
\multicolumn{3}{c}{\textbf{(C) xCOMET-XXL as reference}} \\
\midrule
Comparison & OC Recall & OC Span-F1 \\
\midrule
xCOMET-XXL vs GPT-5.2    & .19 & .26 \\
xCOMET-XXL vs Mistral-Large & .16 & .22 \\
xCOMET-XXL vs GPT-4o-mini & .08 & .13 \\
xCOMET-XXL vs LLaMA-4     & .07 & .11 \\
\bottomrule
\end{tabular}
\caption{Extended OC-based pairwise span agreement on the EU20 human-reference subset. Span-F1 is symmetric across comparison direction, while recall is directional and depends on the chosen reference annotator.}
\label{tab:agreement-oc-extended}
\end{table}

% Table: Logit coefficients (log-odds)
\begin{table}[htpb]
\centering
\tabcolsep=0.26em
\footnotesize
\begin{tabular}{llccccccc}
\toprule
\textbf{Lang} & \textbf{Metric} &
\multicolumn{3}{c}{\textbf{GPT-5.2 vs}} &
\multicolumn{4}{c}{\textbf{Human vs}} \\
\cmidrule(lr){3-5}\cmidrule(lr){6-9}
& &
\textbf{G4o} & \textbf{L4} & \textbf{Mis} &
\textbf{G4o} & \textbf{L4} & \textbf{Mis} & \textbf{GPT-5.2} \\
\midrule
DA & OC  & .23 & .17 & .25 & .12 & .13 & .19 & .48 \\
DE & OC  & .16 & .14 & .19 & .16 & .13 & .20 & .61 \\
ET & OC  & .20 & .20 & .24 & .14 & .14 & .20 & .57 \\
FR & OC  & .17 & .12 & .24 & .10 & .09 & .19 & .34 \\
HU & OC  & .28 & .23 & .28 & .15 & .12 & .22 & .44 \\
IT & OC  & .21 & .18 & .36 & .16 & .14 & .29 & .53 \\
LT & OC  & .23 & .05 & .18 & .16 & .04 & .25 & .49 \\
RO & OC  & .09 & .18 & .25 & .08 & .14 & .21 & .68 \\
SL & OC  & .30 & .26 & .31 & .25 & .23 & .32 & .78 \\
\midrule
DA & SIM & .23 & .08 & .10 & .09 & .08 & .15 & .30 \\
DE & SIM & .14 & .12 & .19 & .16 & .10 & .14 & .39 \\
ET & SIM & .20 & .13 & .13 & .19 & .09 & .15 & .44 \\
FR & SIM & .18 & .13 & .19 & .09 & .06 & .10 & .34 \\
HU & SIM & .16 & .14 & .20 & .10 & .10 & .18 & .49 \\
IT & SIM & .20 & .12 & .24 & .18 & .10 & .25 & .51 \\
LT & SIM & .23 & .03 & .11 & .16 & .02 & .26 & .39 \\
RO & SIM & .11 & .17 & .12 & .09 & .09 & .12 & .42 \\
SL & SIM & .30 & .19 & .24 & .25 & .19 & .22 & .78 \\
\bottomrule
\end{tabular}
\caption{Span-$F_1$ by language for OC and SIM for the four LLM annotators and the human-reference comparison. These values are descriptive rather than inferential, given the small per-language sample size. G4o = GPT-4o-mini, L4 = LLaMA-4, Mis = Mistral-Large.}
\label{tab:f1-by-lang}
\end{table}

\begin{table}[tbhp]
\centering
\tabcolsep=0.68em
\small
\begin{tabular}{lccccc}
\toprule
\textbf{Lang} & \textbf{G4o} & \textbf{L4} & \textbf{Mis} & \textbf{GPT-5.2} & \textbf{Human} \\
\midrule
DA & .00 & .07 & .10 & .00 & .04 \\
DE & .17 & .09 & .20 & .06 & .13 \\
ET & .09 & .09 & .18 & .04 & .05 \\
FR & .10 & .17 & .09 & .07 & .07 \\
HU & .25 & .15 & .14 & .09 & .04 \\
IT & .15 & .22 & .18 & .13 & .13 \\
LT & .00 & .06 & .15 & .04 & .09 \\
RO & .04 & .07 & .12 & .03 & .07 \\
SL & .14 & .29 & .09 & .12 & .11 \\
\bottomrule
\end{tabular}
\caption{Source-overlap rate by language for the four LLM annotators and the human reference (Span-Recall-style micro aggregation): fraction of target error spans (with valid offsets) that overlap target regions linked to GPT-5.2 source-anomaly annotations (OC threshold $0.8$).}
\label{tab:source-overlap-by-lang}
\end{table}

\subsection{Worked Examples for OC and SIM}
\label{sec:app:tqann:metric_examples}

We illustrate the two span-matching criteria from §3 on concrete examples drawn from our annotations.

\paragraph{Example 1: OC with partial overlap.}
Consider a human span $H$ and a candidate match $M$ on the same target text:
\begin{quote}\small
$H$: \textit{heißen Zehentrenner nehmen und ihn auf Ihr Gesicht auftragen} \\
$M$: \textit{einen heißen Zehentrenner nehmen und ihn}
\end{quote}
With target character offsets $H = [811, 871)$ and $M = [805, 845)$, the span lengths and overlap evaluate to
\begin{align*}
|H| &= 60, & |M| &= 40, & |H \cap M| &= 34,
\end{align*}
yielding
\[
\mathrm{OC}(H, M) \;=\; \frac{|H \cap M|}{\min(|H|, |M|)}
\;=\; \frac{34}{40} \;=\; 0.85.
\]
Since $0.85 \geq 0.8$, the spans are matchable under the OC criterion used in the main text. A readable view of the same overlap is:
\begin{quote}\small
Human: \texttt{... einen [H: heißen Zehentrenner nehmen und ihn auf Ihr Gesicht auftragen] ...} \\
Match: \texttt{... [M: einen heißen Zehentrenner nehmen und ihn] auf Ihr Gesicht auftragen ...}
\end{quote}
This example shows that OC can count a genuine partial overlap as a match even when the span boundaries differ substantially.

\paragraph{Example 2: SIM with character-trigram Dice.}
Let $A$ and $B$ be two candidate spans over the same sentence,
\begin{quote}\small
$A$: \textit{sich stärker verfestigen} \\
$B$: \textit{die Mitglieder der Gruppe A sich stärker verfestigen}
\end{quote}
and let $\mathcal{G}_3(s)$ denote the multiset of raw character trigrams of a string $s$. Then $|\mathcal{G}_3(A)| = 22$ and $|\mathcal{G}_3(B)| = 50$. Because $A$ occurs as a contiguous substring of $B$, every trigram in $A$ also occurs in $B$, so the trigram-overlap count equals
\[
\sum_{g} \min\!\bigl(c_g(A),\, c_g(B)\bigr) \;=\; 22,
\]
where $c_g(\cdot)$ is the count of trigram $g$. The Dice similarity is therefore
\[
\mathrm{SIM}(A, B) \;=\; \frac{2 \cdot 22}{22 + 50}
\;=\; \frac{44}{72} \;\approx\; 0.611.
\]
Since $0.611 \geq 0.6$, the two spans are matchable under the SIM criterion. Example shared trigrams include \texttt{sic}, \texttt{ich}, \texttt{stä}, \texttt{ver}, \texttt{tig}, and \texttt{gen}. This example shows that SIM can recover boundary-expanded textual matches even without relying on offsets.

\subsection{Systems-only Multi-Rater Reliability}
\label{sec:app:tqann:reliability}
To make span-level annotation variability explicit, we additionally report systems-only multi-rater reliability on the shared subset using Krippendorff's unitized alpha ($\alpha_u$), which is designed for free-span annotation settings. Because the human reference contains one annotator per language, human--human unitizing agreement is unavailable for this subset. Table~\ref{tab:reliability-main} summarizes the resulting reliability estimates.

\begin{table}[htpb]
\centering
\small
\setlength{\tabcolsep}{18pt}
\renewcommand{\arraystretch}{1.15}
\begin{tabular}{@{}lc@{}}
\toprule
\textbf{Metric} & \textbf{Value} \\
\midrule
Krippendorff $\alpha$                   & 0.037 \\
{\footnotesize\color{gray!70!black}\quad (nominal, any-error)} & \\
Krippendorff $\alpha_u$                 & 0.128 \\
{\footnotesize\color{gray!70!black}\quad (unitizing, target-span localization)} & \\
$\alpha_u$ range across languages        & 0.075--0.186 \\
{\footnotesize\color{gray!70!black}\quad (9 $\times$ 25 segments)} & \\
\bottomrule
\end{tabular}
\caption{Systems-only multi-rater reliability on the shared subset (225 segments, 9 languages, 5 raters). $\alpha_u$ is the global systems-only unitizing reliability for target-text error spans, measuring agreement on span locations and boundaries. The reported range is the min--max of the nine per-language $\alpha_u$ values and is not a confidence interval.}
\label{tab:reliability-main}
\end{table}

\subsection{Threshold-free Character-overlap Metrics}
\label{sec:app:tqann:charoverlap}

To complement thresholded OC/SIM span matching, we also report threshold-free character-level overlap metrics.
Char-F1 is a binary character-overlap measure that ignores severity labels.
Char-F1w is severity-aware and follows a WMT25-style overlap logic, giving full credit to severity-matched overlaps and partial credit (0.5) to severity-mismatched overlaps, with critical grouped with major.
Table~\ref{tab:charoverlap-main} reports the corresponding character-level results, and Table~\ref{tab:threshold-sweep} reports complementary threshold-sweep robustness ranges for OC- and SIM-based matching.

\begin{table}[htpb]
\centering
\footnotesize
\setlength{\tabcolsep}{2pt}
\begin{tabular}{lccc}
\toprule
\textbf{System} & \makecell{\textbf{Any-}\\\textbf{error}\\\textbf{F1}} & \makecell{\textbf{Char-F1w}\\\textbf{{[}95\% CI{]}}} & \makecell{\textbf{Char-F1}\\\textbf{{[}95\% CI{]}}} \\
\midrule
GPT-5.2       & .95 & .42 [.383, .469] & .48 [.437, .529] \\
xCOMET-XXL    & .88 & .20 [.169, .230] & .26 [.223, .301] \\
Mistral-Large & .88 & .21 [.173, .242] & .28 [.235, .317] \\
GPT-4o-mini   & .77 & .17 [.129, .217] & .21 [.163, .261] \\
LLaMA-4       & .63 & .11 [.078, .142] & .14 [.104, .183] \\
\bottomrule
\end{tabular}
\caption{Threshold-free character-overlap results on the shared subset. Any-error F1 is segment-level error/no-error agreement. Char-F1 is threshold-free character-level span-overlap F1 on the binary target-side error mask (global micro, 95\% bootstrap CI via segment resampling). Char-F1w is a severity-aware WMT25-style character-overlap variant with full credit for severity-matched overlaps and partial credit for severity mismatches. Point estimates rounded to two decimals; CI bounds to three.}
\label{tab:charoverlap-main}
\end{table}

\begin{table}[htpb]
\centering
\small
\setlength{\tabcolsep}{18pt}
\renewcommand{\arraystretch}{1.05}
\begin{tabular}{@{}lcc@{}}
\toprule
\textbf{System} & \makecell{\textbf{OC F1}\\\textbf{range}} & \makecell{\textbf{SIM F1}\\\textbf{range}} \\
\midrule
GPT-5.2       & .585--.592 & .426--.506 \\
xCOMET-XXL    & .294--.304 & .154--.229 \\
Mistral-Large & .267--.271 & .157--.208 \\
GPT-4o-mini   & .169--.171 & .122--.148 \\
LLaMA-4       & .138--.144 & .074--.114 \\
\bottomrule
\end{tabular}
\caption{Threshold-sweep robustness on the shared subset. OC/SIM ranges denote min--max micro-F1 across threshold sweeps (OC: 0.7/0.8/0.9; SIM: 0.4/0.5/0.6). Ranges reported at three decimals to preserve the sweep variation.}
\label{tab:threshold-sweep}
\end{table}

\begin{table}[htpb]
\centering
\small
\setlength{\tabcolsep}{10pt}
\renewcommand{\arraystretch}{1.15}
\begin{tabular}{@{}lccc@{}}
\toprule
\textbf{Annotator} & \makecell{\textbf{Spans /}\\\textbf{sample}} & \makecell{\textbf{Median}\\\textbf{span len.}} & \makecell{\textbf{Coverage}\\\textbf{(union)}} \\
\midrule
\textsc{xCOMET}-XXL    & 6.95 & 7.56  & 0.16 \\
Human                  & 6.05 & 9.94  & 0.11 \\
GPT-5.2                & 3.21 & 19.22 & 0.10 \\
Mistral-Large          & 2.74 & 21.17 & 0.11 \\
GPT-4o-mini            & 1.40 & 20.33 & 0.04 \\
LLaMA-4                & 1.29 & 23.17 & 0.04 \\
\bottomrule
\end{tabular}
\caption{Annotator statistics (mean over languages): average number of spans per sample, median span length in characters, and union coverage of target characters. \textsc{xCOMET}-XXL is markedly more fine-grained than the other automatic annotators, which helps explain its relatively higher recall but lower precision under greedy one-to-one matching.}
\label{tab:annotator-stats}
\end{table}

xCOMET-XXL tends to produce more numerous and shorter spans, which helps explain its more recall-oriented behavior under greedy one-to-one matching. Table~\ref{tab:annotator-stats} summarizes these annotator statistics.

\subsection{Bootstrap Confidence Intervals}
\label{sec:app:tqann:bootstrap}

To quantify uncertainty despite the relatively small per-language sample size (25 items per language), we compute pooled stratified bootstrap confidence intervals over items for the four LLM annotators. This auxiliary analysis excludes xCOMET-XXL, since the bootstrap runs were prepared only for the human-vs.-LLM comparisons. In each replicate, we resample 25 items with replacement per language (9 languages; 225 pooled items total), micro-aggregate TP/FP/FN across resampled items, and recompute span-level precision, recall, and F1. We use 2{,}500 bootstrap iterations and report percentile-based 95\% confidence intervals.

Table~\ref{tab:bootstrap-ci-llm} reports the resulting pooled confidence intervals.

\begin{table}[htpb]
\centering
\small
\setlength{\tabcolsep}{12pt}
\renewcommand{\arraystretch}{1.2}
\begin{tabular}{@{}l c c@{}}
\toprule
\makecell{\textbf{Comparison}\\\textbf{Human vs.}} 
  & \makecell{\textbf{OC micro-$F_1$}\\\textbf{{[}95\% CI{]}}} 
  & \makecell{\textbf{SIM micro-$F_1$}\\\textbf{{[}95\% CI{]}}} \\
\midrule
GPT-5.2     & .526 [.484, .568] & .437 [.397, .482] \\
Mistral-Large     & .228 [.198, .259] & .176 [.149, .204] \\
GPT-4o-mini & .141 [.116, .167] & .144 [.119, .171] \\
LLaMA-4     & .120 [.097, .145] & .086 [.066, .107] \\
\bottomrule
\end{tabular}
\caption{Pooled bootstrap confidence intervals for span-level agreement against the human reference, computed for the four LLM annotators. CIs are based on 2{,}500 stratified bootstrap replicates over items (25 items per language, sampled with replacement; 9 languages, 225 pooled items per replicate). Scores are micro-$F_1$ values obtained by aggregating TP/FP/FN across resampled items.}
\label{tab:bootstrap-ci-llm}
\end{table}

\subsection{Additional Qualitative Example Cases}
\label{sec:app:tqann:examples}
To complement the quantitative agreement tables, we include additional target-side (Table~\ref{tab:target-example-cases}) example cases drawn from the manually inspected EU20 subset. The source-side cases (Section~\ref{sec:intro}; Table~\ref{tab:examples_st}) are diagnostic only and are not counted as target-side translation errors. They motivate our separate treatment of source-linked confounds, while the target-side cases illustrate how translation errors can change the construct or answerability of benchmark items.

\begin{table*}[htpb]
\centering
\footnotesize
\setlength{\tabcolsep}{3pt}
\begin{tabularx}{\textwidth}{p{2.5cm} Y Y Y}
\toprule
\textbf{Source} & \textbf{EN snippet} & \textbf{Translation snippet} & \textbf{Distortion type and effect} \\
\midrule
MMLU\,/\,FR \newline \texttt{\scriptsize prof\_medicine/150} &
``A 14-year-old boy\,\ldots\ He is embarrassed because he has grown breasts\,\ldots\ this patient's condition'' &
``\textit{\ldots\ il s'est fait pousser des seins \ldots\ cette patiente}'' &
\textbf{Gender flip / clinical meaning shift:} the male patient becomes female in the translation, changing the diagnostic interpretation from male gynecomastia to female breast development. \\

MMLU\,/\,SL \newline \texttt{\scriptsize moral\_scenarios/349} &
Option (3): ``Not wrong, Wrong''\newline
Option (4): ``Not wrong, Not wrong'' &
Option (3): ``\textit{Ni napačno, ni napačno}''\newline
Option (4): ``\textit{Ni napačno, ni napačno}'' &
\textbf{Structural corruption:} two distinct answer options collapse into the same translation, making the item partially unsolvable. \\

MMLU\,/\,FR \newline \texttt{\scriptsize hs\_world\_hist/84} &
``600 C.E.--1450 C.E.'' &
``\textit{600 à 1450 avant notre ère}'' &
\textbf{Time-period inversion:} C.E.\ becomes B.C.E., shifting the historical scope by roughly two millennia. \\

MMLU\,/\,RO \newline \texttt{\scriptsize prof\_law/115} &
``The estates of the two decedents sued the company'' &
``\textit{Averea celor doi decedați a dat în judecată compania}'' &
\textbf{Legal-semantic shift:} legal \emph{estates} (the juridical entities representing the decedents) becomes personal \emph{wealth/property}, removing the entity that can sue. \\

MMLU\,/\,SL \newline \texttt{\scriptsize hs\_world\_hist/80} &
``in conformity without resolve'' &
``\textit{V skladu z odločitvijo}'' (\emph{with decision}) &
\textbf{Polarity reversal:} the rhetorical meaning flips from ``without resolve'' to ``with decision'', changing the interpretation of the historical statement. \\

MMLU\,/\,SL, ET \newline \texttt{\scriptsize formal\_logic/69, /84} &
``(Cx\,.\,Ox)'' &
``\textit{(Cx\,-\,Ox)}'' &
\textbf{Symbol corruption:} the logical conjunction symbol is replaced by a hyphen, making the formal expression syntactically invalid. \\

MMLU\,/\,DA \newline \texttt{\scriptsize prof\_law/1334} &
``200-acre tract'' &
first ``\textit{200 acres}'', later ``\textit{200 hektar}'' &
\textbf{Measurement distortion:} the translation inconsistently renders acres and then inflates the area to 200 hectares ($\sim$2.5$\times$ larger), altering the legal facts of the case. \\

GSM8K\,/\,DA \newline \texttt{\scriptsize main/test/150} &
``Tim lives 2 miles away from the school'' &
``\textit{Tim bor 3 km væk fra skolen}'' &
\textbf{Number + unit change:} both the distance unit and the quantity are changed, so the translated math problem no longer matches the source. \\

GSM8K\,/\,RO \newline \texttt{\scriptsize main/test/659} &
``A pound of almonds costs \$10 while a pound of walnuts costs \$15'' &
``\textit{Un kilogram de migdale costă 10 dolari \ldots\ un kilogram de nuci costă 15 dolari}'' &
\textbf{Unit distortion:} pound becomes kilogram throughout, changing the price basis and invalidating the original calculation chain. \\

GSM8K\,/\,ET \newline \texttt{\scriptsize main/test/130} &
``Nissa hires 60 seasonal workers to play elves in her department store's Santa village'' &
``\textit{mängivad tontu}'' &
\textbf{Concept mistranslation:} Estonian \emph{tontu} means ghosts/goblins rather than Santa's elves (\emph{päkapikud}), distorting the cultural concept being tested. \\
\bottomrule
\end{tabularx}
\caption{Additional target-side translation cases from our example collection, beyond the four main-text examples. The selected cases illustrate diverse distortion types, including structural corruption, time-period inversion, legal-semantic shifts, polarity reversals, symbol corruption, and unit or concept changes.}
\label{tab:target-example-cases}
\end{table*}

%% file: appendix/tqrefspanaces.tex
\clearpage
\section{Span-ACESRef}
\label{sec:app:tqrefspanaces}

This appendix expands the construction details for \RefSpanACES{} and summarizes its dataset composition and phenomenon-to-MQM mapping.
A key limitation of the original \textsc{Span-ACES} setup is that each item targets exactly one introduced phenomenon, so the released span corresponds only to that targeted error.
If an incorrect translation contains additional issues beyond the targeted one, those extra issues remain unlabeled and can therefore count as false positives during span evaluation.
\RefSpanACES{} reduces this source of noise by projecting the targeted good$\rightarrow$incorrect edit onto the human reference and discarding cases where that projection is not unique or not reliable.

\subsection{Construction, Mapping, and Validation Details}
\label{sec:app:tqrefspanaces:details}

Operationally, we construct \RefSpanACES{} as follows:
\begin{enumerate}
    \item Start from a \textsc{Span-ACES} item consisting of a human reference, a \emph{good} translation, and an \emph{incorrect} translation.
    \item Compute a case-sensitive token diff between the good and incorrect translations.
    \item Keep only items with a single \emph{contentful diff}, i.e., one non-empty contiguous token-level edit that changes lexical content rather than only punctuation or formatting.
    \item Project this edit into the human reference by locating the corresponding reference-side token sequence and use that projected span as the gold span.
    \item Discard items with multiple diffs, no reference-side match, or ambiguous repeated matches in the reference.
\end{enumerate}

This procedure is intentionally conservative: \emph{projection} here does not attempt full semantic alignment, but a simple grounded transfer of the targeted edit into the human reference whenever the match is unique enough to support reliable span evaluation.
In this sense, the original \textsc{Span-ACES} data can be viewed as a targeted one-error-per-item resource, whereas \RefSpanACES{} converts that targeted edit into a cleaner reference-side span benchmark.

\begin{table}[htpb]
\centering
\footnotesize
\setlength{\tabcolsep}{5pt}
\renewcommand{\arraystretch}{1.12}
\begin{tabular}{lrrrr}
\toprule
\textbf{Split} & \textbf{Items} & \textbf{Langs.}
  & \makecell[r]{\textbf{MQM}\\\textbf{types}}
  & \makecell[r]{\textbf{Pheno-}\\\textbf{mena}} \\
\midrule
Accuracy & 1,155 & 20 & 4 & 14 \\
Fluency/Style & 252 & 1 & 1 & 4 \\
\midrule
Total & 1,407 & 20 & 5 & 18 \\
\bottomrule
\end{tabular}
\caption{Final composition of \RefSpanACES{}. The \textit{Fluency/Style} portion is German-only and consists of four anaphoric phenomena; the \textit{Accuracy} portion spans the 20 EU20 target languages.}
\label{tab:span-aces-ref-summary}
\end{table}

Table~\ref{tab:span-aces-ref-summary} summarizes the final dataset composition.
For consistency with our EU20 annotation setup, we map \textsc{Span-ACES} phenomena to MQM types using the ACES authors' released mapping\footnote{\href{https://github.com/EdinburghNLP/ACES}{github.com/EdinburghNLP/ACES}} and collapse the resulting MQM types into two coarse categories: \textit{Accuracy} and \textit{Fluency/Style}.
In total, \RefSpanACES{} contains 1{,}155 \textit{Accuracy} instances spanning the 20 EU20 target languages and 252 \textit{Fluency/Style} instances (German only), for 1{,}407 items overall.
Table~\ref{tab:span-aces-ref-mapping} lists the phenomenon-to-MQM mapping used in \RefSpanACES{}.

\begin{table}[htpb]
\centering
\footnotesize
\setlength{\tabcolsep}{3pt}
\renewcommand{\arraystretch}{1.15}
\begin{tabularx}{\columnwidth}{@{}l X@{}}
\toprule
\textbf{MQM type} & \textbf{ACES phenomenon} \\
\midrule
\multicolumn{2}{@{}l}{\textit{Accuracy}} \\[2pt]
mistranslation & pleonastic\_it:substitution \\
               & coreference-based-on-commonsense \\
               & hallucination-date-time \\
               & overly-literal-vs-ref-word \\
               & overly-literal-vs-explanation \\
               & overly-literal-vs-synonym \\
               & real-world-knowledge-hypernym-vs-distractor \\
               & real-world-knowledge-entailment \\
               & real-world-knowledge-synonym-vs-antonym \\
               & ordering-mismatch \\
untranslated   & untranslated-vs-ref-word \\
               & untranslated-vs-synonym \\
no-translate   & do-not-translate \\
addition       & addition \\
\midrule
\multicolumn{2}{@{}l}{\textit{Fluency/Style}} \\[2pt]
grammar & anaphoric\_intra\_non-subject\_it:substitution \\
        & anaphoric\_intra\_subject\_it:substitution \\
        & anaphoric\_intra\_they:substitution \\
        & anaphoric\_group\_it-they:substitution \\
\bottomrule
\end{tabularx}
\caption{Phenomenon-to-MQM mapping used in \RefSpanACES{}. Accuracy covers 4 MQM types and 14 distinct phenomena; Fluency/Style covers 1 MQM type (\textit{grammar}) and 4 anaphoric phenomena.}
\label{tab:span-aces-ref-mapping}
\end{table}

\begin{table}[htpb]
\centering
\footnotesize
\setlength{\tabcolsep}{4pt}
\renewcommand{\arraystretch}{1.15}
\begin{tabularx}{\columnwidth}{@{}p{2.0cm} X@{}}
\toprule
\textbf{Stage} & \textbf{Illustrative toy example} \\
\midrule
Good translation      & Der Ausschuss \textbf{genehmigte} den Vorschlag. \\
Incorrect translation & Der Ausschuss \textbf{lehnte} den Vorschlag \textbf{ab}. \\
Case-sensitive token diff & Single contentful edit: \texttt{genehmigte} $\rightarrow$ \texttt{lehnte\ldots ab} \\
Human reference       & Der Ausschuss \textbf{genehmigte} den Vorschlag. \\
Projected gold span   & \texttt{genehmigte} \\
Decision              & Keep the item: one contentful diff, unique reference-side match. \\
\bottomrule
\end{tabularx}
\caption{Toy illustration of the good$\rightarrow$incorrect diff and projection step used to derive a gold span in \RefSpanACES{}. The example is schematic and explains the mechanics only. If the human reference did not contain a unique reference-side counterpart of the targeted edit, the item would be discarded.}
\label{tab:span-aces-ref-example}
\end{table}

To sanity-check the transformation pipeline, we manually reviewed a stratified subset of 178 \RefSpanACES{} records using a fixed checklist: (i) whether the projected edit matches the original good$\rightarrow$incorrect change, (ii) whether the intended error type is preserved in the reference context, and (iii) whether the projection introduces side effects elsewhere in the sentence.
We applied the same checklist with GPT-5.2 and compared final \texttt{pass}/\texttt{fail} verdicts.
The two assessments agree on 165/178 items (0.93), indicating that most projected cases are straightforward; the remaining disagreements are mostly borderline cases such as type drift after projection or subtle contextual side effects.

For evaluation, we report both classic and tolerant span metrics.
Classic matching requires exact equality between a predicted span and the gold span.
Tolerant matching is gold-centered and counts a prediction as correct if it contains the gold span and optionally up to $k$ tokens of boundary slack on either side.
We aggregate scores first within each phenomenon and then per MQM category using two weighting schemes: \textbf{mean (N)}, which weights phenomena by sample count, and \textbf{mean (cap)}, which caps each phenomenon at $C{=}25$ to reduce domination by a few large phenomena.

%% file: appendix/tqperf.tex
\clearpage
\section{Performance Analysis}
\label{sec:app:tqperf}

This appendix collects supporting material for Section~\ref{sec:tqperf}.
Section~\ref{sec:app:tqperf:data} summarizes the paired regression dataset,
Section~\ref{sec:app:tqperf:llms} lists the downstream evaluation models,
Section~\ref{sec:app:tqperf:logreg} reports the detailed pooled fixed-effects regressions,
and Section~\ref{sec:app:tqperf:ranking} provides an exploratory counterfactual ranking diagnostic.
Unless noted otherwise, the appendix tables below correspond directly to the confirmatory analyses discussed in the main text.

\subsection{Regression Dataset}
\label{sec:app:tqperf:data}
Table~\ref{tab:tqperf_data_stats} summarizes the paired regression dataset used in Section~\ref{sec:tqperf}.
The paired subset contains 651 unique (item, target-language) pairs and 4{,}993 translated observations across evaluation models after excluding HellaSwag and TruthfulQA.
Besides translated and English correctness counts, we also report how often source-side and target-side annotation lists are empty or non-empty; these annotation counts are over (pair, annotator) records.

\begin{table}[htpb]
\centering
\tabcolsep=0.78em
\footnotesize
\begin{tabular}{lrrrr}
\toprule
\textbf{Statistic} & \textbf{MMLU} & \textbf{GSM8K} & \textbf{ARC} & \textbf{Total} \\
\midrule
$N_{\mathrm{pair}}$ & 237 & 215 & 199 & 651 \\
$N_{\mathrm{obs}}$  & 1896 & 1505 & 1592 & 4993 \\
\midrule
$y{=}1$        & 1127 & 847 & 966 & 2940 \\
$y{=}0$        & 769  & 658 & 626 & 2053 \\
$y^{EN}{=}1$   & 1373 & 1017 & 1255 & 3645 \\
$y^{EN}{=}0$   & 523  & 488  & 337  & 1348 \\
\midrule
$\mathrm{SET}{=}\emptyset$    & 814 & 673 & 722 & 2209 \\
$\mathrm{SET}{\neq}\emptyset$ & 134 & 187 & 74  & 395 \\
$\mathrm{TET}{=}\emptyset$    & 472 & 347 & 478 & 1297 \\
$\mathrm{TET}{\neq}\emptyset$ & 476 & 513 & 318 & 1307 \\
\bottomrule
\end{tabular}
\caption{Dataset statistics for the regression dataset used in Section~\ref{sec:tqperf} (HellaSwag and TruthfulQA excluded). $N_{\mathrm{pair}}$ counts unique (item, target-language) pairs and $N_{\mathrm{obs}}$ translated observations across evaluation models. $y$ and $y^{EN}$ denote translated versus English correctness (counts over $N_{\mathrm{obs}}$). $\mathrm{SET}$/$\mathrm{TET}$ indicate empty vs.~non-empty source-side and target-side annotation lists, respectively; these counts are over (pair, annotator) records.}
\label{tab:tqperf_data_stats}
\end{table}

\subsection{Evaluation Models}
\label{sec:app:tqperf:llms}
Table~\ref{tab:models} lists the eight multilingual instruction-tuned LLMs used as downstream evaluation models in Section~\ref{sec:tqperf}.
We report short names in the main text and the corresponding Hugging Face model identifiers here for reproducibility.
The additional xCOMET-XXL system used in the regression pipeline is a span annotator baseline rather than a downstream evaluation model and is therefore not included in this table.

\begin{table}[htpb]
\centering
\footnotesize
\begin{tabularx}{\columnwidth}{@{}l X r@{}}
\toprule
\textbf{Short name} & \textbf{HF model identifier} & \textbf{\#Params} \\
\midrule
Aya        & \href{https://huggingface.co/CohereLabs/aya-expanse-32b}{\texttt{CohereLabs/aya-expanse-32b}} & 32.3B \\
Command-A  & \href{https://huggingface.co/CohereLabs/c4ai-command-a-03-2025}{\texttt{CohereLabs/c4ai-command-a-03-2025}} & 111B \\
Gemma      & \href{https://huggingface.co/google/gemma-3-27b-it}{\texttt{google/gemma-3-27b-it}} & 27.4B \\
Mistral    & \href{https://huggingface.co/mistralai/Mistral-Small-3.1-24B-Instruct-2503}{\texttt{mistralai/Mistral-Small-3.1-24B-Instruct-2503}} & 24B \\
Pharia     & \href{https://huggingface.co/Aleph-Alpha/Pharia-1-LLM-7B-control-aligned-hf}{\texttt{Aleph-Alpha/Pharia-1-LLM-7B-control-aligned-hf}} & 7.0B \\
Phi        & \href{https://huggingface.co/microsoft/Phi-4-multimodal-instruct}{\texttt{microsoft/Phi-4-multimodal-instruct}} & 5.6B \\
Qwen       & \href{https://huggingface.co/Qwen/Qwen2.5-32B}{\texttt{Qwen/Qwen2.5-32B}} & 32.8B \\
Salamandra & \href{https://huggingface.co/BSC-LT/salamandra-7b-instruct}{\texttt{BSC-LT/salamandra-7b-instruct}} & 7.8B \\
\bottomrule
\end{tabularx}
\caption{Downstream evaluation models used in the performance analysis. The table lists the short names used in the paper, the corresponding Hugging Face model identifiers, and approximate parameter counts.}
\label{tab:models}
\end{table}

\subsection{Detailed Regression Results}
\label{sec:app:tqperf:logreg}
Tables~\ref{tab:tqperf_logit_coef}--\ref{tab:tqperf_ame_specB} report the pooled fixed-effects regressions underlying Section~\ref{sec:tqperf}.
Tables~\ref{tab:tqperf_logit_coef} and \ref{tab:tqperf_logit_or} provide the same four LLM annotator models in log-odds and odds-ratio form, respectively.
The corresponding average marginal effects (AMEs) are the most directly interpretable view and are reported separately per specification:
Table~\ref{tab:tqperf_ame_specA} covers the full model (Spec.~A, with English correctness control), and Table~\ref{tab:tqperf_ame_specB} covers the subset restricted to items solved in English (Spec.~B).
Both AME tables include xCOMET-XXL as an additional span annotator baseline.

% Table: Logit coefficients (log-odds)
\begin{table}[htpb]
\centering
\tabcolsep=0.38em
\footnotesize
\begin{tabular}{@{}l l l l@{}}
\toprule
\textbf{Annotator} & \textbf{Spec.} & \textbf{coef($T$)} & \textbf{coef($S$)} \\
\midrule
GPT-4o-mini & A            & -.40 [-.62, -.19] & -.21 [-.45,  .04] \\
            & $A_{\neg S}$ & -.41 [-.64, -.20] & -- \\
            & B            & -.58 [-.84, -.34] & -.27 [-.55,  .01] \\
            & $B_{\neg S}$ & -.60 [-.86, -.36] & -- \\
\midrule
GPT-5.2     & A            & -.41 [-.64, -.19] & -.21 [-.45,  .03] \\
            & $A_{\neg S}$ & -.42 [-.66, -.20] & -- \\
            & B            & -.56 [-.86, -.31] & -.28 [-.57, -.01] \\
            & $B_{\neg S}$ & -.57 [-.86, -.31] & -- \\
\midrule
LLaMA-4     & A            & -.44 [-.72, -.16] & -.21 [-.45,  .04] \\
            & $A_{\neg S}$ & -.45 [-.73, -.20] & -- \\
            & B            & -.61 [-.95, -.29] & -.25 [-.54,  .03] \\
            & $B_{\neg S}$ & -.64 [-.97, -.34] & -- \\
\midrule
Mistral-Large & A          & -.33 [-.58, -.10] & -.22 [-.47,  .03] \\
               & $A_{\neg S}$ & -.35 [-.58, -.12] & -- \\
               & B            & -.38 [-.66, -.12] & -.29 [-.58, -.01] \\
               & $B_{\neg S}$ & -.40 [-.69, -.13] & -- \\
\bottomrule
\end{tabular}
\caption{Logit coefficients (log-odds) for translation errors $T$ and source-side issues $S$, with 95\% block-bootstrap confidence intervals. Spec.~A includes English correctness $y^{EN}$; Spec.~B restricts to $y^{EN}{=}1$. $A_{\neg S}$ and $B_{\neg S}$ omit $S$. All models include fixed effects for target language, dataset, and evaluation model.}
\label{tab:tqperf_logit_coef}
\end{table}

% Table: AME, Spec. B (English-correct subset)
\begin{table}[htpb]
\centering
\footnotesize
\setlength{\tabcolsep}{8pt}
\renewcommand{\arraystretch}{1.15}
\begin{tabular}{@{}l l c c@{}}
\toprule
\textbf{Annotator} & \textbf{Spec.} & \textbf{AME($T$)} & \textbf{AME($S$)} \\
\midrule
GPT-4o-mini   & B            & \makecell{-9.64\\\scriptsize[-14.05, -5.49]}  & \makecell{-4.34\\\scriptsize[-8.75, 0.19]} \\
              & $B_{\neg S}$ & \makecell{-10.01\\\scriptsize[-14.52, -5.86]} & -- \\
\midrule
GPT-5.2       & B            & \makecell{-8.98\\\scriptsize[-13.35, -4.93]}  & \makecell{-4.64\\\scriptsize[-9.19, -0.10]} \\
              & $B_{\neg S}$ & \makecell{-9.26\\\scriptsize[-13.46, -5.03]}  & -- \\
\midrule
LLaMA-4       & B            & \makecell{-10.71\\\scriptsize[-16.56, -4.91]} & \makecell{-4.15\\\scriptsize[-8.66, 0.42]} \\
              & $B_{\neg S}$ & \makecell{-11.23\\\scriptsize[-17.10, -5.57]} & -- \\
\midrule
Mistral-Large & B            & \makecell{-6.12\\\scriptsize[-10.30, -1.86]}  & \makecell{-4.74\\\scriptsize[-9.29, -0.20]} \\
              & $B_{\neg S}$ & \makecell{-6.48\\\scriptsize[-10.74, -2.18]}  & -- \\
\midrule
xCOMET-XXL    & B            & \makecell{-6.20\\\scriptsize[-10.88, -1.57]}  & \makecell{-4.91\\\scriptsize[-9.45, -0.51]} \\
              & $B_{\neg S}$ & \makecell{-6.48\\\scriptsize[-11.06, -1.77]}  & -- \\
\bottomrule
\end{tabular}
\caption{Average marginal effects (AMEs; probability points) of target-side translation errors $T$ and source-side issues $S$ on correctness in translation under \textbf{Spec.~B} (subset with $y^{EN}{=}1$, i.e., items the model solves correctly in English). $B_{\neg S}$ omits $S$. Point estimates are shown with 95\% block-bootstrap CIs below. All models include fixed effects for target language, dataset, and evaluation model.}
\label{tab:tqperf_ame_specB}
\end{table}

% Table: Odds ratios
\begin{table}[htpb]
\centering
\tabcolsep=0.40em
\footnotesize
\begin{tabular}{@{}l l l l@{}}
\toprule
\textbf{Annotator} & \textbf{Spec.} & \textbf{OR($T$)} & \textbf{OR($S$)} \\
\midrule
GPT-4o-mini & A            & .67 [.54, .83] & .81 [.64, 1.04] \\
            & $A_{\neg S}$ & .66 [.53, .82] & -- \\
            & B            & .56 [.43, .71] & .77 [.58, 1.01] \\
            & $B_{\neg S}$ & .55 [.42, .70] & -- \\
\midrule
GPT-5.2     & A            & .67 [.53, .83] & .81 [.64, 1.03] \\
            & $A_{\neg S}$ & .66 [.52, .82] & -- \\
            & B            & .57 [.43, .74] & .75 [.56,  .99] \\
            & $B_{\neg S}$ & .56 [.43, .73] & -- \\
\midrule
LLaMA-4     & A            & .65 [.49, .85] & .81 [.63, 1.04] \\
            & $A_{\neg S}$ & .63 [.48, .82] & -- \\
            & B            & .54 [.39, .75] & .78 [.58, 1.03] \\
            & $B_{\neg S}$ & .53 [.38, .71] & -- \\
\midrule
Mistral-Large & A          & .72 [.56, .90] & .80 [.63, 1.03] \\
               & $A_{\neg S}$ & .71 [.56, .88] & -- \\
               & B            & .69 [.52, .89] & .75 [.56,  .99] \\
               & $B_{\neg S}$ & .67 [.50, .88] & -- \\
\bottomrule
\end{tabular}
\caption{Odds ratios (OR) for translation errors $T$ and source-side issues $S$, with 95\% block-bootstrap confidence intervals. Spec.~A includes English correctness $y^{EN}$; Spec.~B restricts to $y^{EN}{=}1$. $A_{\neg S}$ and $B_{\neg S}$ omit $S$. All models include fixed effects for target language, dataset, and evaluation model.}
\label{tab:tqperf_logit_or}
\end{table}

% Table: AME, Spec. A (full model, with EN control)
\begin{table}[htpb]
\centering
\footnotesize
\setlength{\tabcolsep}{8pt}
\renewcommand{\arraystretch}{1.15}
\begin{tabular}{@{}l l c c@{}}
\toprule
\textbf{Annotator} & \textbf{Spec.} & \textbf{AME($T$)} & \textbf{AME($S$)} \\
\midrule
GPT-4o-mini   & A            & \makecell{-6.74\\\scriptsize[-10.44, -3.17]} & \makecell{-3.50\\\scriptsize[-7.46, 0.61]} \\
              & $A_{\neg S}$ & \makecell{-6.99\\\scriptsize[-10.62, -3.22]} & -- \\
\midrule
GPT-5.2       & A            & \makecell{-6.76\\\scriptsize[-10.61, -3.11]} & \makecell{-3.54\\\scriptsize[-7.47, 0.57]} \\
              & $A_{\neg S}$ & \makecell{-7.01\\\scriptsize[-10.80, -3.26]} & -- \\
\midrule
LLaMA-4       & A            & \makecell{-7.51\\\scriptsize[-12.47, -2.70]} & \makecell{-3.47\\\scriptsize[-7.56, 0.62]} \\
              & $A_{\neg S}$ & \makecell{-7.83\\\scriptsize[-12.50, -3.33]} & -- \\
\midrule
Mistral-Large & A            & \makecell{-5.59\\\scriptsize[-9.50, -1.69]}  & \makecell{-3.69\\\scriptsize[-7.66, 0.44]} \\
              & $A_{\neg S}$ & \makecell{-5.82\\\scriptsize[-9.47, -2.04]}  & -- \\
\midrule
xCOMET-XXL    & A            & \makecell{-4.60\\\scriptsize[-9.07, 0.00]}   & \makecell{-3.82\\\scriptsize[-7.69, 0.19]} \\
              & $A_{\neg S}$ & \makecell{-4.82\\\scriptsize[-9.10, -0.44]}  & -- \\
\bottomrule
\end{tabular}
\caption{Average marginal effects (AMEs; probability points) of target-side translation errors $T$ and source-side issues $S$ on correctness in translation under \textbf{Spec.~A} (full model, controlling for English correctness $y^{EN}$). $A_{\neg S}$ omits $S$. Point estimates are shown with 95\% block-bootstrap CIs below. All models include fixed effects for target language, dataset, and evaluation model.}
\label{tab:tqperf_ame_specA}
\end{table}

\subsection{Counterfactual Ranking Analysis}
\label{sec:app:tqperf:ranking}
This subsection reports an exploratory transparency analysis for the question of whether translation errors affect only absolute translated scores or also downstream leaderboard orderings.
Using the fitted regression models, we predict each evaluation model's translated accuracy under the counterfactual setting $T{=}0$ (no target-side translation errors) and compare the resulting ranking to the observed ranking.
Table~\ref{tab:tqperf_rankcorr} reports bootstrapped rank correlations for three representative annotators (Mistral-Large, LLaMA-4, and GPT-5.2), and Table~\ref{tab:tqperf_rankuplift} reports the per-model counterfactual uplift $\Delta$, defined as the predicted increase in translated accuracy (in percentage points) when moving from the observed setting to $T{=}0$.

\begin{table}[htpb]
\centering
\small
\setlength{\tabcolsep}{6pt}
\renewcommand{\arraystretch}{1.12}
\begin{tabular}{@{}lccc@{}}
\toprule
\textbf{Metric} & \textbf{Mistral-Large} & \textbf{LLaMA-4} & \textbf{GPT-5.2} \\
\midrule
Spearman $\rho$ & 0.9992 & 0.9995 & 0.9995 \\
95\% CI         & [0.9762, 1.0] & [1.0, 1.0] & [1.0, 1.0] \\
Kendall $\tau$  & 0.9976 & 0.9984 & 0.9986 \\
95\% CI         & [0.9286, 1.0] & [1.0, 1.0] & [1.0, 1.0] \\
Boot            & 2500 / 0 & 2500 / 0 & 2500 / 0 \\
\bottomrule
\end{tabular}
\caption{Rank correlation between observed evaluation-model rankings and counterfactual rankings predicted under $T{=}0$. Correlations are bootstrapped over underlying items ($B=2500$). ``Boot'' reports successful / failed bootstrap runs. Near-perfect correlations indicate that the relative ordering of evaluation models is highly stable under the $T{=}0$ intervention in this paired subset.}
\label{tab:tqperf_rankcorr}
\end{table}

\begin{table}[htpb]
\centering
\footnotesize
\setlength{\tabcolsep}{2pt}
\renewcommand{\arraystretch}{1.0}
\begin{tabular}{@{}lcc@{}}
\toprule
\multicolumn{3}{c}{\textbf{(a) Annotator = Mistral-Large}} \\
\midrule
\textbf{Eval model} & \textbf{$\Delta$ mean} & \textbf{95\% CI} \\
\midrule
Phi-4-multimodal-instr.-6B          & 4.57 & [1.44, 7.71] \\
aya-expanse-32b-Instr.             & 4.39 & [1.39, 7.37] \\
Salamandra-7b-instr.                & 4.29 & [1.32, 7.33] \\
Qwen-2.5-32B-Instr.                 & 4.13 & [1.30, 6.96] \\
Mistral-Small-3.1-24B-Instr.-2503   & 4.10 & [1.29, 6.86] \\
c4ai-command-a-03-111B-Instr.  & 3.85 & [1.21, 6.44] \\
Pharia-1-LLM-7B-contr.-aligned       & 3.84 & [1.18, 6.59] \\
Gemma-3-27B-Instr.                  & 3.59 & [1.12, 6.01] \\
\midrule
\multicolumn{3}{c}{\textbf{(b) Annotator = LLaMA-4}} \\
\midrule
\textbf{Eval model} & \textbf{$\Delta$ mean} & \textbf{95\% CI} \\
\midrule
Qwen-2.5-32B-Instr.                 & 1.36 & [-0.21, 2.91] \\
Phi-4-multimodal-instr.-6B          & 1.27 & [-0.20, 2.73] \\
aya-expanse-32b-Instr.              & 1.27 & [-0.19, 2.73] \\
Mistral-Small-3.1-24B-Instr.-2503   & 1.20 & [-0.18, 2.60] \\
Salamandra-7b-instr.                & 1.18 & [-0.19, 2.50] \\
c4ai-command-a-03-111B-Instr.  & 1.14 & [-0.17, 2.46] \\
Gemma-3-27B-Instr.                  & 1.09 & [-0.16, 2.36] \\
Pharia-1-LLM-7B-contr.-aligned       & 1.06 & [-0.17, 2.24] \\
\midrule
\multicolumn{3}{c}{\textbf{(c) Annotator = GPT-5.2}} \\
\midrule
\textbf{Eval model} & \textbf{$\Delta$ mean} & \textbf{95\% CI} \\
\midrule
Phi-4-multimodal-instr.-6B          & 2.45 & [-0.46, 5.44] \\
aya-expanse-32b-Instr.              & 2.36 & [-0.45, 5.22] \\
Salamandra-7b-instr.                & 2.25 & [-0.42, 5.03] \\
Qwen-2.5-32B-Instr.                 & 2.25 & [-0.42, 4.96] \\
Mistral-Small-3.1-24B-Instr.-2503   & 2.22 & [-0.42, 4.89] \\
c4ai-command-a-03-111B-Instr.  & 2.09 & [-0.39, 4.55] \\
Pharia-1-LLM-7B-contr.-aligned       & 2.01 & [-0.38, 4.48] \\
Gemma-3-27B-Instr.                 & 1.95 & [-0.37, 4.32] \\
\bottomrule
\end{tabular}
\caption{Per-model counterfactual uplift $\Delta$ (percentage points), defined as the predicted increase in translated accuracy when moving from the observed setting to the counterfactual setting $T{=}0$. Confidence intervals are bootstrapped over underlying items ($B=2500$). Although the ranking changes are negligible, the corresponding score uplifts remain non-trivial, indicating that translation errors still bias absolute multilingual performance estimates downward.}
\label{tab:tqperf_rankuplift}
\end{table}

Taken together, the near-perfect rank correlations and non-zero score uplifts suggest that, in this paired subset, translation errors behave mostly like an approximately shared downward shift across systems: relative leaderboard orderings are largely preserved, but absolute translated scores remain biased downward.
This is why we treat ranking stability as a useful diagnostic rather than as evidence that translation quality is unimportant.

%\subsection{Note on Finer-Grained Slices}
%\label{sec:app:tqperf:exploratory-note}
%Earlier drafts also examined finer-grained per-language and model$\times$language slices.
%In the paired source-controlled subset, however, these cells become small quickly, especially after conditioning on English correctness or splitting by annotator, which makes such estimates unstable and difficult to interpret.
%For the camera-ready paper, we therefore restrict the appendix to the pooled fixed-effects regressions above and the explicit ranking diagnostic in Section~\ref{sec:app:tqperf:ranking}.
%Any finer-grained slices should be treated as descriptive exploratory material rather than confirmatory evidence.